\documentclass[letterpaper]{article} % DO NOT CHANGE THIS
\usepackage{aaai23}  % DO NOT CHANGE THIS
\usepackage{times}  % DO NOT CHANGE THIS
\usepackage{helvet}  % DO NOT CHANGE THIS
\usepackage{courier}  % DO NOT CHANGE THIS
\usepackage[hyphens]{url}  % DO NOT CHANGE THIS
\usepackage{graphicx} % DO NOT CHANGE THIS
\usepackage{natbib}  % DO NOT CHANGE THIS AND DO NOT ADD ANY OPTIONS TO IT
\usepackage{caption} % DO NOT CHANGE THIS AND DO NOT ADD ANY OPTIONS TO IT
\usepackage{algorithm}
\usepackage{algorithmic}
\usepackage{bm}
\usepackage{amsfonts}  
\usepackage{amsmath}
\usepackage{nicefrac}
\usepackage{graphics}
\usepackage{graphicx}
\usepackage{booktabs}
\usepackage{amsthm}

\providecommand{\expect}{\bm{E}}

\providecommand{\citep}{\cite} 
\providecommand{\citet}{\cite}

\usepackage{newfloat}
\usepackage{listings}
\DeclareCaptionStyle{ruled}{labelfont=normalfont,labelsep=colon,strut=off} % DO NOT CHANGE THIS
\floatstyle{ruled}
\newfloat{listing}{tb}{lst}{}
\floatname{listing}{Listing}
\newtheorem*{theorem*}{Theorem}
\newtheorem{theorem}{Theorem}
\newtheorem{proposition}{Proposition}

\newtheorem{lemma}{Lemma}
\newtheorem{assumption}{Assumption}
\usepackage{amsfonts}
\title{Learning Optimal Features via Partial Invariance}
\author{
    {Moulik Choraria\textsuperscript{\rm 1,}\footnote{corresponding author}, Ibtihal Ferwana\textsuperscript{\rm 1}, Ankur Mani\textsuperscript{\rm 2}, Lav R. Varshney\textsuperscript{\rm 1}}
}
\affiliations{
\textsuperscript{\rm 1}University of Illinois at Urbana-Champaign, \\ \textsuperscript{\rm 2} University of Minnesota Twin Cities\\

\{moulikc2, iferwna2, varshney\}@illinois.edu, amani@umn.edu
}

\usepackage{bibentry}
\begin{document}

\maketitle

\begin{abstract}
Learning models that are robust to distribution shifts is a key concern in the context of their real-life applicability. Invariant Risk Minimization (IRM) is a popular framework that aims to learn robust models from multiple environments. The success of IRM requires an important assumption: the underlying causal mechanisms/features remain invariant across environments. When not satisfied, we show that IRM can over-constrain the predictor and to remedy this, we propose a relaxation via \emph{partial invariance}. In this work, we theoretically highlight the sub-optimality of IRM and then demonstrate how learning from a partition of training domains can help improve invariant models. Several experiments, conducted both in linear settings as well as with deep neural networks on tasks over both language and image data, allow us to verify our conclusions. 
\end{abstract}

\section{Introduction}
Standard machine learning models trained using classical Empirical Risk Minimization (ERM) can be expected to generalize well to unseen data drawn from the same distribution as the training data \cite{vapnik2013}. However, distribution shifts during test time (when data is from different sources or under different conditions) can severely degrade model performance \cite{beeryGP2018, lakeRTGS2017, marcus2018}.  The errors can often be attributed to the model picking up statistically informative but spurious correlations, which in turn limits their real-life applications since in practice, the use-case almost always differs from the training data. Thus, several lines of research explore alternate learning objectives for training robust models.\\
One particular line of research stems from the Invariant Causal Prediction framework \cite{jonasPBM2015}, where the goal is to learn causal mechanisms that work well under interventions; our work focuses on the similarly inspired Invariant Risk Minimization (IRM) framework, which aims to learn a predictor that relies only on features that are invariant across all training environments. The underlying motivation for invariance is rooted in its strong links with causality \cite{judeaP2009}, with the intuition being that by invariance can help the model distinguish the causal features from domain-specific spurious features, which it can then discard for better generalization. \\
%A standard assumption in such invariance-based objectives is sufficiency \cite{ahuja_erm_irm2020}, in that there exists a predictor relying solely on the set of invariant features that achieves optimal risk in all environments.  Hence, in settings where the invariant features are insufficient or settings with \textit{concept drift}, where the conditional distribution of the label w.r.t.\ the causal features can itself change, it is unclear whether the IRM framework (and other similar objectives) can achieve the desired performance. However, such situations often arise in practical applications, for instance in discussion forums, when although some common linguistic features are shared among all communities, other features might have different connotations within different communities \cite{gallacher2021,ManiV2021}. In such situations however, IRM (or similar objectives) is often directly applied to the entire set of available data/training environments, without accounting for these factors \cite{maximePGJAPCKW2021, robertACMZ2020}, for instance in language tasks where distribution shifts have been noted across time \cite{luu2021time_embeddings} or comment toxicity classification, where accounting for the community context is crucial. Therefore in practice, imposing invariance constraints across all environments may end up over-constraining the model and cause performance to degrade, especially when such non-invariant yet informative features are prone to being discarded by the model.\\
A standard assumption in such invariance-based objectives is that of sufficiency \cite{ahuja_erm_irm2020}: there exists a predictor, relying solely on the invariant features, which can achieve Bayes optimal risk in all environments. While fairly general, this assumption may not be satisfied for certain classes of distribution shifts. For instance, consider a prediction task with \textit{concept drift}, wherein the relationship of the `causal' features (features that are causally responsible to the label) with the label changes across training environments. Here, a predictor relying solely on invariant features ends up being over-constrained, since it is incentivized to discard non-invariant but informative causal features that are needed for Bayes optimality. Such situations are ubiquitous in practice, for instance in language tasks in which linguistic features can have different connotations within different communities \cite{gallacher2021,ManiV2021} or in tasks with distribution shifts across time \cite{luu2021time_embeddings}. Additionally, even when a sufficient representation exists theoretically, it may not be accessible due to shortcomings in the optimization of the learning objective. However, these factors are seldom accounted for when considering the application of IRM (or other invariant learning objectives) for a given use-case \cite{maximePGJAPCKW2021, robertACMZ2020}. Thus, it is important to develop a characterization for the same. \\
To address this gap in literature, we present a first study to characterize the behaviour of IRM under explicit concept drifts. Then, we take a step further and propose a relaxation for IRM via the \emph{Partial Invariance} (P-IRM) framework. We find that our framework increases the flexibility of invariant models by allowing learning of features that are locally invariant within a partition of the training environments. This flexibility is accompanied with an inherent trade-off; the cost of finding the right partition, in an information-agnostic setting, grows exponentially in the number of environments. However, for certain classes of problems, including the language tasks alluded to previously, readily available meta-information often allows us to easily infer the `optimal' training partition for a given use-case. Notice that in doing so, we move away from the OoD minimax regime, and instead focus on improving generalization conditioned on availabity of this meta-information. In this work, we begin by presenting a theoretical characterization of IRM under concept shifts. Next, we formally quantify the notion of meta-information and assuming access to it, we theoretically and empirically demonstrate how the notion of Partial Invariance can help improve the performance of invariant models. The rest paper is organized as follows: we begin with a literature review in Sec.~\ref{sec:rw}, and motivate P-IRM and present our main results in Sec.~\ref{sec:theory}. We report our empirical evaluations in Sec.~\ref{sec:experiments} and wrap up with some concluding remarks in Sec.~\ref{sec:discussion}. 

\section{Related Work}
\label{sec:rw}

Many approaches aim to learn deep invariant feature representations: some focus on domain adaptation by finding a representation whose distribution is invariant across source and target distributions \cite{shaiBCKPV2010, kunGS2015}, while others focus on conditional domain-invariance  \cite{gongZLTGS2016, liGTLT2018}. However, there is evidence that domain adaption approaches are insufficient when the test distribution may lie outside the convex hull of training distributions \cite{leeR2018,duchiGN2018, mohriSS2019}. Other approaches include Bayesian Deep Learning \cite{nealR1996}, which tries to account for model uncertainty during test-time, and Robust Optimization \cite{benTEN2009}, which aims to generalize well to distributions close to training.\\
Our work focuses particularly on the IRM framework \cite{arjovskyBGP2020}, which relates to domain generalization wherein access to the test distribution is not assumed. IRM is rooted in the theory of causality \cite{scholkop2012} and proposes invariance for achieving OoD generalization \cite{petersBM2016,heinzeCM2018}. In \cite{ahujaSVD2020}, the authors reformulate IRM via a game-theoretic approach, wherein the invariant representation corresponds to the Nash equilibrium of a game. While the IRM framework assumes only the invariance of the conditional expectation of the label given the representation, some follow-ups rely on stronger invariance assumptions \cite{xieYCLSL2021, mahajanTS2021}. As mentioned before, this line of work assumes sufficiency of invariant features whereas we specifically focus on distribution shifts when sufficiency is violated.\\
Several follow-up works attempt to characterize IRM's performance under different settings and model assumptions. It has been noted that carefully tuned ERM can often outperform state-of-the-art domain generalization approaches, including IRM, across multiple benchmarks \cite{gulrajaniL2020}. The failure of IRM may stem from the gap between the proposed framework and its practical ``linear'' version (IRMv1), which fails to capture natural invariances \cite{KamathTDS2021}. Indeed, the authors of \cite{rosenfeldRP2020} demonstrate that a near-optimal solution to the IRMv1 objective, which matches IRM on training environments, does no better than ERM on environments that differ significantly from training. Following these deficiencies, several works propose alternate objectives for achieving invariance \cite{kruegerCJ2021, alexisBV2020, wengongJBJ2020, kartikACZGBMR2021, ChangjianSWG2021}. \\
However, unlike previous works that aim to improve the invariance learning objective, we question whether invariance as a constraint can be improved upon for better performance. To that end, our notion of \emph{partial invariance} generalizes not only IRM, but all similar invariance learning objectives. The use of meta-information for invariant learning has been proposed in \cite{lin2022zin}. However, unlike partitioning, the focus therein is to artificially generate environment membership for samples when not available a priori. Finally, a related idea appears in \cite{RunpengYZLHZYHH2022}, which proposes applying different invariance penalty weights for different domains, but with the goal of addressing data quality variance across domains.  
\section{Theory}
\label{sec:theory}
In this section, we present the notion of partial invariance.\\
\textbf{Notation:} We use upper-case boldface $\bm{U}$ to denote matrix/tensor/vector valued random variables, and lowercase boldface $\bm{u}$ to denote scalar valued random variables. We use upper-case $U$ to denote matrices/vectors/tensors and lowercase $u$ to denote scalars.

\subsection{Invariant Risk Minimization}
The IRM setup assumes access to datasets of the form $D_e:=\{X^e_i, y^e_i\}_{i=1}^{n_e}$ collected from multiple training environments $e \in \mathcal{E}_{tr}$. The samples in dataset $D_e$ are i.i.d.\ from the environment's joint distribution, $P(\bm{X}^e, \bm{y}^e)$. The task is to estimate a map $f: \mathcal{X} \rightarrow \mathcal{Y}$ or alternatively, the conditional distribution $P(Y|X)$, so that it performs well across unseen environments $\mathcal{E}_{all} \supset \mathcal{E}_{tr}$. Formally, the IRM framework aims to minimize the Out-of-Distribution (OoD) risk: $R^{OoD}(f) = \max_{e \in \mathcal{E}_{all}} R^{e}(f)$, where $R^{e}(f) := \expect_{\bm{X}^e, \bm{y}^e}[\ell (f(\bm{X}^e), \bm{y}^e)]$ is the expected risk in environment $e$. {The predictor $f$ is parametrized as $w \circ \Phi$, wherein $\Phi: \mathcal{X} \rightarrow \mathcal{Z}$ represents the learned representation and $w: \mathcal{Z} \rightarrow \mathcal{Y}$ is a linear predictor over said representation. The IRM learning objective is posed as a constrained optimization problem:
}
\begin{align}
&\min_{\Phi, w} \sum_{e\in \mathcal{E}_{obs}} {R}^e(w \circ \Phi) \tag{\textbf{IRM}} \\ 
\mbox{s.t.} & \ w \in \text{arg}\min_{\tilde{w}} {R}^{e'}(\tilde{w} \circ \Phi)\ \forall\ {e'} \in \mathcal{E}_{tr} \label{eq_irm}\mbox{.}
\end{align}
To avoid the inner optimization, the minimization constraint is replaced by a more tractable gradient penalty: 
\begin{align}
&\min_{\Phi, w} \sum_{e\in \mathcal{E}_{obs}} {R}^e(w \circ \Phi) \tag{\textbf{IRMgc}}\\ 
\mbox{s.t.} & \ w \in \{{\tilde{w}}: \|\nabla_{w}{R}^{e'}(w \circ \Phi)\|=0\ \forall\ {e'} \in \mathcal{E}_{tr}\} \label{eq_irmgc} \mbox{,}
\end{align}
where IRMgc is shorthand for the gradient constrained IRM. In practice, this constraint is enforced via a regularizer $\lambda$:
\begin{align}
&\min_{\Phi} \sum_{e\in \mathcal{E}_{obs}} {R}^e(\Phi) + \lambda \|\nabla_{w, w=1.0}{R}^{e}(\Phi)\| \mbox{,} \tag{\textbf{IRMv1}}  \label{eq_irm_v1} 
\end{align}
where the implicit overparametrization in having a separate classifier and representation map is removed by fixing a dummy classifier $w=1.0$. Thus, $\Phi$ becomes the entire invariant predictor and the strictness of the gradient norm penalty, which enforces invariance, is via $\lambda$. Note that when $\lambda=\infty$, IRMv1 is equivalent to IRMgc, which in turn is the first order approximation for the true IRM objective.\\  
An intrinsic assumption in the IRM learning setup for proving minimax optimality is the ideal scenario of sufficiency i.e. there exists a $\Phi$ that is invariant across all $e \in \mathcal{E}_{tr}$ and is sufficient i.e $\bm{y}^e \perp \bm{X}^e |\ \Phi(\bm{X}^e)\ \forall e \in \mathcal{E}_{all}$ \cite{ahuja_erm_irm2020}. However if sufficiency is violated for an environment, one would expect the IRM model, which relies solely on invariant features, to be sub-optimal for that environment (compared to a model that utilizes non-invariant features along with invariant ones). Such a situation may arise under \emph{concept drift}, wherein the the conditional expectation of the label $\bm{y}_e$ given ``causal'' features may change across environments. Thus, in practice, if an invariant $\Phi$ that is also sufficient for  environments does not exist for the desired use-case, we expect the performance of IRM (or related frameworks) to degrade. We illustrate this with a simple example.\\
\textbf{Example 1} We adapt the generative model from \cite{arjovskyBGP2020}:  the goal is to predict target $y$ using $X = [x_1, x_2, x_3]$, in environment $e$ such that $e \in \mathcal{E}_{all}$ can affect the distribution of $X$ as well as the conditional distribution of $y$ given $X$ via {a deterministic map $c(e): \mathcal{E}_{all}\rightarrow \{-1, 1\}$}:
%probabilistic map $c(e)$:
 %\vspace{-1mm}
\begin{equation*}
    %\vspace{-1mm}
    \begin{gathered}
     \bm{x}_1 \leftarrow N(0, \sigma(e)^2), \bm{x}_2 \leftarrow N(0, \sigma(e)^2), \\
     %c(e) \leftarrow \text{Unif}(\{1, -1\}), \epsilon \sim N(0, \sigma(e)^2)\\
     {c(e) \in \{1, -1\}, \epsilon \sim N(0, \sigma(e)^2)}\\
     \bm{y} \leftarrow \bm{x}_1 + c(e)\bm{x}_2 + \epsilon,\ \epsilon \perp \bm{x}_1, \ \epsilon \perp \bm{x}_2\\
     \bm{x}_3 \leftarrow \bm{y} + N(0, 1), \sigma(e)^2 \in [0,  \sigma_{MAX}^2].\\
    \end{gathered}
\end{equation*}
We estimate $\bm{y}$ as $\hat{\bm{y}} = \alpha_1 \bm{x}_1 + \alpha_2 \bm{x}_2 + \alpha_3 \bm{x}_3$. Within the IRM framework, the only feasible representation $\Phi$ (upto scaling) that yields invariant predictors across all $e$ is $\Phi([\bm{x}_1, \bm{x}_2, \bm{x}_3]) = [\bm{x}_1, 0, 0]$, with corresponding regression coefficients $[1, 0, 0]$. Although this minimizes the OoD error for arbitrary $e$, it does so by discarding the non-invariant but informative $\bm{x}_2$. However, if our predictor is privy to some knowledge of $c(e)$, we could first partition the set of training environments $\mathcal{E}_{tr}$ into two partitions, such that environments within a partition have the same $c(e)$ value. Then, applying IRM within each partition yields models with better performance that can exploit $\bm{x}_2$ as an invariant feature in the partition. Note that this partial notion of invariance still retains the ability to discard spurious/non-causal $\bm{x}_3$. Additionally with partitioning, we can improve generalization if information about $c(e_{unseen})$ is available, by choosing the right model/partition for prediction. Next, we study the conditions under which partitioning can improve upon IRM performance and we refer to this method as P-IRM.

\subsection{Model}

 For our analysis, we consider a simple regression task to succinctly capture our intuition about the conditions under which partitioning is feasible. To begin with, we assume access to the underlying causal features and instead, focus on understanding the nature of the IRM solution set under distribution shifts. In the next part, we extend this analysis to study learning under partial invariance.\\
 We consider the following generative model: we observe samples $(X_i^e, y_i^e)$in environment $e$ and the goal is to predict $y_i^e$ from $X_i^e$. $X_i^e$'s are samples corresponding to the random variable $\bm{X}^e \sim P(\bm{X}^e)$, as described below:
%$$\bm{x}^e = [x_1^e, x_2^e, \ldots, x_c^e]^\top \sim N(\bm{\mu(e)},
%\bm{\Sigma}(e)),\ \bm{x}^e \in \mathbb{R}^c,$$
$$\bm{X}^e = [\bm{x}_1^e, \bm{x}_2^e, \ldots, \bm{x}_c^e]^\top \sim P(\bm{X}^e),$$
%\bm{\Sigma}(e)),\ \bm{x}^e \in \mathbb{R}^c,
 %We consider the set of features to be subdivided into levels, $\bm{x}^i$ represents the feature at level $i$ and the features are independent (or disentangled) from each other.
 %\textbf{Assumption 1} Each individual feature is centered i.e. $\bm{\mu_i}(e) = \bm{0}\ \forall\ (i, e)$.\\
 where each $\bm{x}_i^e$ denotes an individual feature. To simplify our initial analysis, we assume that the individual features are independent of each other and are normalized i.e. $\expect[\bm{X}^e] = \bm{0}$ and $\expect[\bm{X}^e {\bm{X}^e}^\top] = \bm{I}\ \forall\ e$. The target $\bm{y}^e$ for given $\bm{X}^e$ can be characterized as:
 \begin{align}
 \begin{split}
  & \bm{y}^e =  \langle {W^e}, \bm{X}^e \rangle + \epsilon_y \mbox{,}\\
  & W^e = [w_1^e, w_2^e, \ldots, w_c^e] \in \mathbb{R}^c, \epsilon_y \sim N(0, \sigma_y^2(e)).
  \end{split}
 \end{align} 
 where weights $W^e$ encode the conditional distribution of observing label $\bm{y}^e$ given $\bm{X}^e$ in environment $e$ and are fixed for that environment, and $\langle \cdot , \cdot \rangle$ denotes the standard inner product in $\mathbb{R}^c$. 
 %The generative model describes the causal relationship $X \rightarrow Y$, with $\expect[Y^e|\bm{X}^e = \bm{X}] = \langle {\bm{W}^e}, \bm{X} \rangle$. For a given level, the set of possible weights for the feature at level $i$ is $A_i$. Thus, an environment is characterized by a unique selection of weights $\bm{W}^e$ across levels, as:
 For a given feature $\bm{x}_i^e$ in environment $e$, the corresponding feature weight $w_i^e$ is independently and uniformly sampled from set $A_i$ for each environment $e$. Once sampled however, these weights remain fixed for that environment. Additionally $|A_1| = 1$, so that feature weight $w_1^e$ is fixed and thus $\bm{x}_1^e$ is invariant for all $e$:  
 \begin{equation}
    \label{eq:gen_model}
    \begin{gathered}
    \bm{W}^e = [\bm{w}_1^e, \bm{w}_2^e, \ldots, \bm{w}_c^e], \\
    \text{ where } \ {\bm{w}_{i}}^e \sim \text{Unif}(\{A_i\})\ \forall\ i \in \{1, 2, \ldots ,c\}, \\ |A_1|=1, \ |A_i| > 1\ \forall\ i > 1.
    \end{gathered}
\end{equation}
%\textbf{Assumption 2} In addition to the above model,  for a given $i \in {1, 2,\ldots,L}$, for all $\bm{a}_{i,k} \in A_i$, we have that $\bm{\|v}\|_2 = m_i$ and that $m_i > m_j$ for all $(i < j)$.\\
%We make note of some important aspects of the described model. Firstly, environments are drawn uniformly from a countable $\mathcal{E}_{all}$, determined by the cardinality of sets $\{A_i\}_{i=1}^L$. We introduce a notion of levels ($i \in {1, 2,\ldots,L}$) with respect to the feature weights, noting that the feature weight at level $i$ is uniformly sampled from set $A_i$, whose cardinality increases as level $i$ increases. Then, we note that $\bm{x_1}$ is the only \emph{truly invariant feature} since $\expect[Y^e|\bm{x_1}] = \bm{a_{inv}}^\top \bm{x_1}$, is fixed for all environments $e \in \mathcal{E}_{all}$. Additionally, the cardinality of $A_i,\ |A_i| = c_i$ captures the degree of invariance of feature $x_i$, with a higher cardinality indicating the conditional expectation is more likely to change across environments. Our assumption on feature weight magnitudes introduces a notion of feature importance, since feature weights that are more likely to vary across environments (higher $c_i$) have a lower magnitude $m_i$.\\
We make note of some important aspects. As per our model, ${\bm{x}_1^e}$ is the only \emph{truly invariant feature} since $\expect[\bm{y}^e|{\bm{x}_1^e}] = w_{inv}.\bm{x}_1^e$, is fixed for all $e$, where $A_1 = \{w_{inv}\}$ is a singleton, and $w_{inv}$ denotes the invariant feature weight. Additionally, the cardinality of set $A_i, |A_i|$ defines an implicit notion of the variance of feature $x_i$, with a higher cardinality indicating that the feature weight is more likely to change across environments and is thus, \emph{less invariant}.\\
 %We consider the task of predicting $y^e$ given $\bm{x}^e$, under the mean squared loss. Recall that the IRM framework considers predictors of the form $\bm{w}\circ \bm{\Phi}$, where the transformation $\bm{\Phi} \in \mathbb{R}^{c \times c}$ extracts a subset of features $\Phi \bm{X}^e$ from $\{\bm{x}_i\}_{i=1}^L$. Notice that to recover feature $\bm{x}_i$ at level $i$, we need the $i$th row of $\bm{\Phi}$, $\bm{\Phi}_{i,\cdot} = \delta_i^T$, where $\delta_i \in \mathbb{R}^L$ denotes the standard basis vector. Thus, the identity matrix $\bm{\Phi} = I_{L} \in \mathbb{R}^{L \times L}$ recovers all features from $\bm{X}$. We can then apply weights $\bm{W} = [\bm{w_1}, \bm{w_2}, \ldots, \bm{w_L}]^T$, $\bm{W} \in \mathbb{R}^{L \times c}$ to obtain the label prediction as $\hat{Y} = \langle \bm{W},\bm{\Phi}\bm{X}\rangle$. Then, the expected error of using a given predictor $\hat{Y} = \bm{W} \circ \bm{\Phi}$ in an environment $e$ equates to:
 % \begin{equation}
 % \label{eq:mse_risk}
 %     E_{\bm{X}_e, Y_e }[({Y^e} - \hat{Y}(\bm{X}^e))^2] = E_{\bm{X}_e, \epsilon_y}[(\langle \bm{W^e}, \bm{X^e} \rangle + \epsilon_y - \langle \bm{W},\bm{\Phi} \bm{X^e} \rangle)^2].
 % \end{equation}
 With our generative model in place, we next consider the task of predicting $\bm{y}^e$ given $\bm{X}^e$, under the mean squared loss. Recall that the IRM framework considers predictors of the form $w \circ \Phi$, where the transformation $\Phi$ extracts a suitable representation and $w$ is the linear predictor acting on that representation. Due to the implicit overparametrization, we fix $w=1.0$ to a scalar value as proposed in \cite{arjovskyBGP2020} and analyze the corresponding IRM solutions with ${\Phi} \in \mathbb{R}^c$. For simplicity, we ignore finite sample effects and consider the objective in \eqref{eq_irm_v1} when $\lambda = \infty$, or equivalently, the gradient penalty constraint in \eqref{eq_irmgc} which ideally approximates the true IRM objective. Additionally, we assume the following for training environments $\mathcal{E}_{tr}$.
 \begin{assumption}[\emph{Sufficiency for IRM}] \label{assump:irm} Assume $\exists$ an environment $e\in \mathcal{E}_{tr}$ for which the truly invariant predictor is sufficient, i.e. the corresponding feature weights satisfy $w_1^e = w_{inv}$ and ${w^e}_i = 0\ \forall\ i \in \{2, \ldots, c\}$.
 \end{assumption} \noindent In other words, we assume existence of a training environment in which the invariant predictor that only recovers the invariant feature $\bm{x}_1^e$ achieves optimal MSE risk, which is a standard assumption in related literature \cite{ahuja_erm_irm2020}.  
 \begin{lemma}
   As per above parametrization (with $w=1.0$), under Assumption \ref{assump:irm}, the values for ${\Phi}$ that satisfies the IRM solution constraints in \eqref{eq_irmgc} is a singleton and the value of the corresponding predictor equates to ${\Phi} = [w_{inv}, 0, 0\ldots, 0],$ the predictor only recovers the invariant feature.        \label{lemma:1} 
 \end{lemma}
 \noindent
 The proof of the lemma is included in the Appendix and relies on showing that any predictor which assigns non-zero weights to any of non-invariant features would violate the gradient penalty constraints. More importantly, the previous lemma roughly says that any non-invariant feature will be discarded by the IRM predictor. Note that while this is a desirable property for minimax optimality, we ask whether we can do better given additional contextual information. \\
 We formalize the notion of contextual information explicitly by defining an oracle $\omega(e) = \mathbf{1}[ \|W^{e_{ref}} - W^{e}\|_0 \leq \delta]$, that provides us a notion of distance between environments, from a fixed reference environment $e_{ref}$. Alternatively, it identifies whether environment $e$ is close to $e^{ref}$.\\ 
 \textbf{Remark 1}: The choice of the $\ell_0$ metric for the oracle suits our combinatorial setting, since we do make any assumptions on the individual elements in the feature weight sets (i.e. $A_i$'s).\\
 Next we characterize our objective to utilize this information. Suppose we know that our test environment shares the feature weight with the reference environment for a given feature $\bm{x}_i^e$. Then we can define the goal of minimizing the risk w.r.t. to the predictor $f$, conditioned on this information: 
 $$R^{cond}(f) = \expect_{e \mbox{ s.t. } w^{e_{ref}}_i = w^{e}_i} R^{e}(f),$$ 
 where the expectation is over the draw of environments as per the uniform sampling.
  We note that a predictor that accounts for the prior condition (reference feature) will improve performance (i.e. with a lower MSE risk $R^{cond}$), as compared to the truly invariant predictor in the previous lemma. However, to obtain the required feature as a feasible solution via IRM constraints, we need to first isolate a subset of training environments $\mathcal{E}_{partition}\subseteq \mathcal{E}_{tr}$ such that within this set, $w^e_i$ is invariant and secondly, that we avoid learning the rest of the non-invariant features to avoid feature weight mismatches in unseen environments. It turns out that with access to the oracle and under certain mild conditions, we can ensure exactly that in our uniform distribution shift model. Before stating the result, we require a similar sufficiency assumption for the \emph{partially invariant predictor}. 
  \begin{assumption}[Sufficiency for P-IRM]
  \label{assump:pirm}
  Assume $\exists$ an environment $e\in \mathcal{E}_{tr}$ for which the partially invariant predictor is sufficient, i.e. the corresponding feature weights satisfy $w_1^e = w_{inv}$, $w_i^e = w_i^{e_{ref}}$ and ${w^e}_j = 0\ \forall\ j \in \{2, \ldots, c\}\setminus \{i\}$.   
  \end{assumption}
 \begin{theorem}
 \label{thm:1}
 Under the model \eqref{eq:gen_model}, under Assumption \ref{assump:pirm}, with access to oracle $\omega(e) = \mathbf{1}[ \|W^{e_{ref}} - W^{e}\|_0 \leq \delta]$ and $\delta < (c-2)/2$, isolate $\mathcal{E}_{partition} := \{e \in \mathcal{E}_{tr}| \omega(e) = 1\} \cup \{e_{ref}\} \subseteq \mathcal{E}_{tr}$. Next, let $|A_i| = k$, where $A_i$ is the set corresponding to the feature weight $w_i^{e_{ref}}$ of interest. Then, if the sets $\{A_j\} \forall\ j \in \{2, \ldots, c\} \setminus \ \{i\}$ satisfy $|A_j| > \alpha k$ for some $\alpha > 1$, %and as before, in the new set of training environments  $\mathcal{E}_{partition}$, $\exists (e, e')\in \mathcal{E}_{partition}$ such that for the weight for each feature $j \in \{2, \ldots, c\}\setminus \ \{i\}$, satisfies ${w^e}_j < {w^{e'}}_j$, %satisfies that $|A_{j}^{partition}| \geq 2\  \forall\ j \in \{2, \ldots, c\} \setminus \ \{i\}$, 
 we have with probability greater or equal to $(\frac{p}{p+1})^{|\mathcal{E}_{partition}|}$, where $p \geq \frac{(c-1 - \delta)\alpha}{\delta}$, the IRM solution over set $\mathcal{E}_{partition}$ will recover the feature of interest $w_i^{e_{ref}}$.  
 \end{theorem}
 %define a representation $\bm{\Phi}$ that recovers the feature $\bm{x}_i$ (for some $i > 1$) during training, with feature weights $\bm{w}_{i, e_{train}}$, improves upon the invariant predictor iff:
 %$$ 2\bigg(1-\tfrac{1}{m_i^2(c_i-1)} {\sum}_{\substack{\bm{v} \in A_i \\ \bm{v} \neq \bm{w}_{i, e_{train}}}} \langle \bm{v}, \bm{w}_{i, e_{train}} \rangle_{2} \bigg) \leq \tfrac{c_i}{c_i-1} = \tfrac{1}{P(\bm{w}_{i, e_{test}} \neq \bm{w}_{i, e_{train}})}, $$
 %where $\langle \cdot, \cdot \rangle_2$ denotes the standard inner product in $\mathbb{R}^c$.
\noindent The proof, available in the Appendix, relies on showing that within the partition that satisfies the oracle condition, the probability of successfully isolating the required feature is high. Then the result follows as a consequence of Lemma \ref{lemma:1}.\\
In words, the theorem says that if we can identify a partition in which the environments are not too different, then with high probability, the IRM solution will recover features which do not vary too much (i.e. non-invariant but still close to invariant). Note that in 
case of erroneous partitioning, the solution set allowed by the non-convex penalty becomes harder to characterize due to the presence of other feature weights besides the reference. Nevertheless, if the conditions are such that probability of that happening is sufficiently low, we can safely assume that partitioning will achieve a better expected risk. Additionally, it suggests that P-IRM becomes feasible as the oracle becomes more precise and the feature of interest is close to invariant. \\
\textbf{Remark 2}: While P-IRM does improve upon the IRM solution, both variants are likely to be outperformed by ERM in this setting. However, we point out that this is a simplified setting wherein access to causal features is assumed. In more general settings when the causal features need to be inferred from complex data, ERM may be susceptible invariance to confounders/anti-causal variables and thus, we require invariance as a means to make the solution robust. %And while not universally, we verify that P-IRM can indeed outperform both ERM and IRM empirically in certain settings. 
%allowing us to isolate the error for each level $i$. Under the uniform sampling of environments, with a large feature weight set cardinality $|A_i|= c_i$ that is sufficiently non-degenerate (for distinct $\bm{u}, \bm{v} \in A_i$, the correlation between $\bm{u}, \bm{v}$ is low), the optimal representation will almost always correspond to the truly invariant representation. Alternatively, the optimal $\bm{\Phi}$ will only choose features that are close to being invariant (small $c_i$ and strongly correlated feature weights in $A_i$). However, if additional context allows us to deduce that $P(\bm{w}_{i,e_{train}} \neq \bm{w}_{i, e_{test}})$ is small, the requisite condition for retaining feature $x_i$ at level $i$ is satisfied, meaning that we can indeed improve upon the invariant predictor.
\subsection{Partitioning and Partial Invariance}
 Next, we study P-IRM in a general setup, using previous results to characterize the required number of training environments as in IRM. As before, we assume access to the oracle, $\omega$ to identify the partition, i.e. $\mathcal{E}_{partition} \subseteq \mathcal{E}_{tr}$. \\ 
% \begin{figure}
%     \centering
%     \begin{subfigure}{0.48\textwidth}
%     \centering\captionsetup{width=.9\linewidth}
%     \includegraphics[width=0.9\linewidth]{figures/tree_envs_a.png}
%     \caption{An environment is represented as a path between nodes (denoting choice of feature weight) at different levels.}
%     \end{subfigure}
%     \begin{subfigure}{0.48\textwidth}
%     \centering\captionsetup{width=.9\linewidth}
%     \includegraphics[width=0.9\linewidth]{figures/tree_envs_b.png}
%     \caption{A partition is denoted by a subset of nodes (in nodes) across levels, and all paths through this subset represents environments in that partition.}
%     \end{subfigure}
%     \caption{Illustration of an environment and a partition in our model. An observed environment (a path) is denoted by a solid line, while unobserved ones are dotted. The environments in the partition are colored red, whereas those not in the partition are black.}
%     \label{fig:tree_model}
% \end{figure}
\textbf{Learning Setup}: We consider the same causal mechanism for regression task $(\bm{x}^e, y^e)$ from before. The goal is to find a partition using the oracle such that a feature of interest corresponding to the reference environment, $w_i^{e_{ref}}$ is retained. Note that since we want to retain only the invariant features denoted as ${\bm{X}_{inv}}^e = [\bm{x}_1^e, \bm{x}_i^e]$, and discard the non-invariant (or non-partially invariant) features, we encapsulate them into the noise term as $\tilde{\epsilon}_y = {\epsilon}_y + (\bm{X}^{e}_{\{1 \cdots c\} \setminus \{1, i\}})^\top W^{e}_{\{1 \cdots c\} \setminus \{1, i\}}$. Then, notice that we still have $\tilde{\epsilon}_y \perp {\bm{X}_{inv}}^e$ and that $\expect[\tilde{\epsilon}_y] = 0$, due to feature independence and centering assumptions. Next, we consider a realistic learning setup where we observe a scrambled version $\tilde{\bm{X}}^e$ of the true causal features ${\bm{X}^e}$:
\begin{equation}
\label{eq:observe_model}
\begin{gathered}
    \bm{y}^e = ({\bm{X}_{inv}}^e)^\top {W_{inv}}^e + {\tilde{\epsilon}}_y,\ \  \tilde{\epsilon}_y \perp {\bm{X}_{inv}}^e, \ \  \expect[\tilde{\epsilon}_y] = 0\\
    \tilde{\bm{X}}^e = S(\bm{X}^e, \bm{X'}^e).
\end{gathered}
\end{equation}
 Here, $\bm{X}^e = [{\bm{X}_{inv}}^e, \bm{X}^{e}_{\{1 \cdots c\} \setminus \{1, i\}}] \in \mathbb{R}^{c}$ denote the causal features with respect to the label, $\bm{X'}^e \in \mathbb{R}^{q}$, $\tilde{\bm{X}}^e = S(\bm{x}^e, \bm{d}^e) \in \mathbb{R}^{d}$ with $S \in \mathbb{R}^{d \times (c+q)}$. The variable $\bm{X'}^e$ may be arbitrarily correlated with ${\bm{X}_{inv}}^e, \tilde{\epsilon}_y$ or the label $\bm{y}^e$ and is intended to represent the spurious correlations in data. However, we require $S$ to be such that $\exists \tilde{S}$ s.t. $\tilde{S}(S(\bm{X}^e, \bm{X'}^e)) = {\bm{X}_{inv}}^e$ i.e. an inverse map such that the recovery of the desired features is feasible.\\ 
 Next, we define $\gamma = \frac{1}{k\sqrt{2n}}\exp(-n D(\delta/n \| 1/\alpha k)),$ where as before, $\delta$ is the oracle distance parameter, $k$ is the cardinality of the set $A_i$, $|A_i|$, $\alpha$ is as defined in Theorem \ref{thm:1}, $n = c-2$ and $D(m\|n)$ denotes KL divergence between $\mbox{Bern}(m)$ and $\mbox{Bern}(n)$. %Herein, $\gamma$ estimates the lower bound on the probability of sampling an environment that satisfies both the oracle requirement and contains the required feature. Then we have the following sample complexity on the number of required environments.
 Intuitively, $\gamma$ estimates the lower bound on the probability of sampling an environment under the generative model that satisfies the oracle condition of close distance to the reference environment. Then we have the following sample complexity on the number of required environments.

 \begin{theorem}[Informal]
 \label{thm:2}
Assume we observe $(\tilde{\bm{X}}^e, \bm{y}^e)$ as per \eqref{eq:observe_model}, with environments $e \in \mathcal{E}_{tr}$ sampled as per \eqref{eq:gen_model} and let $\mathcal{E}_{partition} := \{e \in \mathcal{E}_{tr}| \omega(e) = 1\} \cup e_{ref} \subseteq$. Let $\bm{\Phi} \in \mathbb{R}^{d \times d}$ have rank $r > 0$. Then sampling $|\mathcal{E}_{tr}| >  \frac{1}{\gamma}(d-r+d/r)\log(1/\epsilon)$ ensures partition cardinality $|\mathcal{E}_{partition}| > d - r + d/r$ with probability $> 1-\epsilon$. Furthermore, if $e \in \mathcal{E}_{partition}$ lie in linear general position of degree $r$ (Assumption 3 in Appendix), then with probability greater than or equal to $(\frac{p}{p+1})^{|\mathcal{E}_{partition}|}$, where $p \geq \frac{(c-1 - \delta)\alpha}{\delta}$, the oracle identifies $\mathcal{E}_{partition}$ such that the predictor $w \circ \Phi$ learnt via IRM within that partition recovers the desired features/weights and corresponding prediction $({\bm{X}}_{inv}^e)^\top {W}_{inv}^e$,  $\forall e \in \mathcal{E}_{all}$ which satisfy $w^{e}_i = w^{e_{ref}}_i$.

 \end{theorem}

\noindent The proof along with the formal statement is included in the Appendix and follows from our previous results by applying concentration bounds on the draw of environments, and subsequently using prior generalization results for IRM. In words, Theorem~\ref{thm:2} states that if the obtained partition is accurate, is of sufficient cardinality and is sufficiently diverse, then ${\Phi}$ recovers the partially invariant features. However, notice that the required number of environments grows inversely with $\gamma$, meaning that we need stronger priors (i.e. sample environments close to the reference) to obtain feasible sample complexities in the number of required environments. 
\subsection{Partial Invariance in Practice}
Next, we state the P-IRM objective more formally. We first assume a distance metric $d$ between environments (known directly or via contextual information). Then, our goal is to identify a subset of training environments $\mathcal{E}_{partition} \subseteq \mathcal{E}_{tr}$ such that its average distance w.r.t. a reference environment $e^{ref}$ roughly satisfies: 
$$ \frac{1}{|\mathcal{E}_{partition}|} \sum_{e \in \mathcal{E}_{partition}} d(e, e^{ref}) < \frac{1}{|\mathcal{E}_{tr}|} \sum_{e\in\mathcal{E}_{tr}} d(e, e^{ref}).$$
%In practice, we rarely have access to the underlying causal features that allow for clustering. Additionally, the number of training environments are far fewer than those required in Theorem $~\ref{thm:2}$ to allow for meaningful clusters. However, access to contextual information can allow us to estimate a notion of similarity between environments (in terms of the underlying causal features), thus allowing us to understand the topology of the underlying graph. 
Thus, the predictor is trained on a subset of observed environments. However, discarding environments is not data-efficient and can lead to lower fidelity and worse generalization, especially in high-complexity models. To avoid this, we introduce the notion of \textit{conditional invariance} as an alternative. Formally, consider the set of observed training environments $\mathcal{E}_{tr}$ and a subset corresponding to the partition $\mathcal{E}_{partition}$ (chosen suitably via $d$), satisfying $\mathcal{E}_{partition} \subseteq \mathcal{E}_{tr}$. We propose the following two variants of P-IRM:
\begin{align*}
&\min_{\Phi, w} \sum_{e\in \mathcal{E}_{1}} {R}_e(w \circ \Phi) \mbox{ s.t.}\ w \in \text{arg}\min_{\tilde{w}} {R}_{e'}(\tilde{w} \circ \Phi)\ \forall\ {e'} \in \mathcal{E}_{2} \mbox{,} \\
&\ \text{if } \mathcal{E}_{1} = \mathcal{E}_{2} = \mathcal{E}_{partition} \tag{\textbf{P-IRM (Partitioning)}},\\
&\ \text{if } \mathcal{E}_{1} = \mathcal{E}_{tr}\ \text{\&} \  \mathcal{E}_{2} = \mathcal{E}_{partition} \tag{\textbf{P-IRM (Conditioning)}}
\end{align*}
%\akhil{Parenthesis not matched for equation tags}
where the empirical risk minimization objective is over environments in $\mathcal{E}_{1}$ and the IRM invariance constraint is applied on environments in $\mathcal{E}_{2}$. For P-IRM (Conditioning), note that while the model uses data from all environments, the invariance penalty is applied only to environments within the chosen partition, which mitigates the issue of having fewer data samples. Intuitively, it serves as a relaxation of the IRM objective to allow for partially invariant features. 
Next, we qualitatively discuss some potential issues in the application of P-IRM. Firstly, fulfilling the requirements as per Theorem \ref{thm:2}, for the required worst case number of environments is infeasible. Fortunately, in practice, IRM can pick up the required invariances from just two environments and we expect P-IRM to overcome that issue as well. \\
Next, we revisit the idea of the distance oracle. While a precise characterization of the distance between causal features of different domains is essentially unobtainable in practice, certain situations allow for inferring the nature of the distribution shift via available contextual information which, while often discarded by practictioners, can serve as an effective pseudo-metric for the same. For instance, authors of \cite{luu2021time_embeddings} pointed out that temporal mis-alignments of distributions in language tasks leads to performance degradation, noting that degradation increases with an increase in the time duration between test and train environments. Thus, learning from only the recent past could yield a larger and more relevant set of invariant features for a use-case on future data. %Intuitively, one can imagine concept shift over time as the true feature weight vector evolving over time but with certain regularity properties, so that environments closer in time provide a better estimation for the future weights. \\

\section{Experiments}
\label{sec:experiments}
We start with a basic sanity check via a synthetic experiment, as an extension of the example presented earlier to visualize how IRM can end up suppressing non-invariant causal features, leading to performance degradation. We then evaluate the efficacy of the P-IRM framework (both partitioning or conditioning) on four tasks: a regression task for housing price prediction, an image classification task on the MetaShift dataset \cite{liang2022metashift}, an entity recognition task for scientific texts on the SciERC dataset \cite{luan2018sciERCdataset} dataset, and a text classification task for prediction of venues of scientific papers. Within image classification, we consider two sub-tasks: Domain Generalization and Sub-population shifts. We defer the synthetic experiment on IRM, along with the text classification and Sub-population shift tasks to the Appendix. \\ 
 For baselines besides IRM, we evaluate the results for standard ERM as well Information Bottleneck IRM (IB\_IRM) \cite{kartikACZGBMR2021}. In addition, we include experiments in the image and language tasks to empirically characterize the effect of partitioning on ERM and IB\_IRM, which we dub as P-ERM and P-IB\_IRM respectively.\\
%\textbf{Remark 3}: We emphasize that partitioned alis implemented in exactly the same manner as their standard counterparts, with the only difference being that the training set is a chosen subset of all available training data. We find that learning from less data via partitioning can be beneficial not just for IRM, but for ERM and other invariant learning algorithms as well.\\ 
An underlying thread for our experiments is the availability to meta-information that allows us to estimate a notion of distance or similarity between environments, which P-IRM can then exploit to construct the required partitions. Specifically, in both housing price prediction and entity recognition task, our environments are partitioned across time and due to distribution shifts, we expect environments closer in time to have higher similarity. Similarly in MetaShift, meta-labels for each image is made available within the data-set, that allows an explicit notion of the distance between training and testing environments. In all our experiments we employ the train-domain validation strategy\cite{ishaanGP2020} for hyper-parameter tuning. The code is available at \url{https://github.com/IbtihalFerwana/pirm} and other implementation details are deferred to Appendix. %However, we do not tune for the IRMv1 penalty parameter, fixing it at $\lambda = 10^3$ to ensure it is large enough to simulate true IRM. %{\ankur{Each experiment has been carefully chosen to demonstrate different effects. Summarize the effects explored in the experiments.}}

\subsection{Linear Regression} 
We consider a regression task to predict house prices based on house features \footnote{House Prices Dataset: \url{https://www.kaggle.com/c/house-prices-advanced-regression-techniques}}, built across years [1910-2010]. Each data point consists of 79 predictive features (for instance, number of bedrooms or house area) and a corresponding target, which is the house price. As pre-processing, we drop all non-numerical features, samples with missing values and normalize each feature and price labels to zero mean, unit variance with the samples, $\{X_i, y_i\}_{i} \in (\mathbb{R}^{32} \times \mathbb{R})$.\\
\textbf{Experiment Setup} To adapt this task to OoD prediction, following \cite{lin2022zin}, we manually split the training data-set into 10-year segments and use the house year built as a meta-data for partitioning, with the intuition being that factors affecting house prices change over time.\\
For prediction, we consider a linear regression model for the  task. Since the IRM framework $w \circ {\Phi}$ is inherently overparametrized, we fix $w=1.0 \in \mathbb{R}$ and we consider $\Phi \in \mathbb{R}^{32}$ (prediction ($\Phi^\top \bm{X}$)) with the Adam optimizer \cite{diederikKB2015}. We consider 6 training environments corresponding to years [1910-1970], while the test samples draw from 4 OoD environments [1970-2010]. We expect partitions closer to the test set to yield better predictors.\\
\textbf{Results} We report the test MSE error (both average and worst group) over the set of testing OoD environments, averaged over $5$ random seeds in Table 
\ref{tab:housing_results}. We find that P-IRM significantly improves the average and worst group OoD error over IRM. Partitioning also benefits ERM, showing more evidence of a distribution shift over time, %as in Fig. ~\ref{fig:housing_erm_vs_irm} in Appendix.
evidence presented in the Appendix. Finally, note that for the two variants for P-IRM, partitioning performs much better where we have more samples than parameters.  

\begin{table}[!ht]
\centering
\small
\tabcolsep=0.08cm
    % \scalebox{0.80}{
  \begin{tabular}{c|ccc}
  \toprule
  Model & Training & Avg. MSE & Worst Group MSE \\
  \midrule
  ERM & 1910-1970 &  0.475 (0.000) & 1.037 (0.000) \\
  ERM & 1930-1970 & 0.431 (0.000)  & 0.963 (0.000) \\
  IRM &  1910-1970  & 0.522 (0.015)  & 1.129 (0.038) \\
  P-IRM (partitioned) & 1930-1970 & \textbf{0.427 (0.009)} & \textbf{0.873 (0.024)} \\
  P-IRM (conditioned) & 1930-1970 & 0.490 (0.014) & 1.035 (0.034) \\
  \bottomrule
  \end{tabular}%
  % }
%\vspace{6pt}
\normalsize
\caption{House Prices Shifts, partitioning demonstrates improvement for both ERM and IRM, test set is 4 OoD environments consisting of houses built between 1970-2010.}
\label{tab:housing_results}

\end{table}%

\subsection{Image Classification}
\label{sec:metashift_section}
We evaluate P-IRM on a binary image classification task on the MetaShift dataset \cite{liang2022metashift}. \\ \textbf{Dataset} In MetaShift dataset, each image is associated with a set of tags that describe the image context (e.g., cat on a \textit{rug}, cat beside a \textit{chair}). Thus, for each given tag (e.g.\ \textit{rug}, \textit{chair}), there is an associated set of images and these sets can overlap if an image has multiple tags. This structure naturally induces a graph, with each image context $C_i$ denotes a node (or \textit{community}) in the graph. This graph is weighted and the weights between nodes is determined by the number of images that are shared between the communities. The weights between each pair of communities, $C_i$ and $C_j$, estimate the similarity between two communities and are calculated using the  Szymkiewicz-Simpson coefficient, which yields the corresponding adjacency matrix $\bm{G}$:  
\begin{equation}
    G(i,j) = \tfrac{|C_i\cap C_j|}{\min(|C_i|,|C_j|)}
    \label{eq:overlap}
\end{equation}
Having access to such an undirected weighted graph over sets of images thus allows us to derive an implicit notion of distance between the corresponding communities.\\ \textbf{Notion of Distance} To introduce partitioning, we develop a notion of distance, which then allows us to quantify the relatedness between training and testing environments. These environments are assumed to be sets of communities. To estimate the distance $d$ between any two given nodes/communities, given that our data is structured as a weighted graph, we can make use of the \textit{spectral embeddings} \cite{spectralEmbeddings}. Spectral embeddings are based on graph Laplacian connectivity \cite{ng2001spectral}. The graph Laplacian $\bm{L}$ is calculated by $\bm{L} = \bm{D}_{diag}-\bm{G}$, where $\bm{D}_{diag}$ is a diagonal degree matrix of the graph $\bm{G}$. The corresponding eigenvectors of $L$, $\bm{u}_1, \dots, \bm{u}_k$, computed and normalized to form the matrix $\bm{U}$, are the corresponding embeddings for the graph. Once we calculate the spectral embeddings, we measure $d$ between communities as the euclidean distance between the corresponding spectral embeddings of each community node. With our notion of distance, we can partition the graph based on distances between sets of communities and identify subsets of training communities which are closer to the test environment. \\ \textbf{Experiment Setup} For all our experiments, we consider the same set of training communities as in ~\cite{liang2022metashift}, which are split into two environments in the IRM setting. To introduce partitioning, we assume distances $d$ between the training environments and the test communities is known/can be estimated via the meta-labels. For learning the P-IRM model, we consider the training environment for IRM which is closer to the test set on average, and split it into two sub-environments. Note that under this split, P-IRM has access to roughly only half the training samples compared to IRM. To remedy this, we consider additional data splits wherein we add samples from communities in the other IRM training environment, that are close to the test set. These additional samples amount to a percentage $p$ of samples in that environment, allowing P-IRM access to a slightly larger portion of the training set. Following \cite{liang2022metashift}, we fix the test community to be $dog(shelf)$ and vary distance $d$ between dog train vs test communities. The cat training set remains unchanged. \\   % Table generated by Excel2LaTeX from sheet 'better_looking'
  \begin{table*}[!ht]
  \small
    \centering
    \tabcolsep=0.03cm
      \begin{tabular}{cccccc}
      \toprule
      %     &
      %     \multicolumn{1}{l}{Experiment 1} & \multicolumn{1}{l}{Experiment 2} & \multicolumn{1}{l}{Experiment 3} & \multicolumn{1}{l}{Experiment 4} &  \\
      % \midrule
          & $d=0.17$ & $d=0.54$ & $d=0.81$ & $d=0.92$ & Avg. Performance \\
      \midrule
      ERM & 0.777(0.078) & 0.560(0.179) & 0.493(0.119) & \textbf{0.667(0.114)} & 0.62425 \\
      P-ERM $(p=0)$ & \textbf{0.823(0.045)} & \textbf{0.790(0.086)} & 0.387(0.074) & 0.663(0.192) & 0.66575 \\
      P-ERM $(p=10)$ & 0.820(0.098) & 0.770(0.057) & 0.493(0.141) & 0.663(0.128) & \textbf{0.6865} \\
      % P-ERM $(p=25)$ & \textbf{0.867(0.050)} & 0.740(0.079) & \textbf{0.557(0.056)} & 0.430(0.079) & 0.6485 \\
      \midrule
      IRM & 0.757(0.231) & 0.477(0.172) & \textbf{0.757(0.110)} & 0.687(0.309) & 0.6695 \\
      P-IRM $(p=0)$ & \textbf{0.960(0.050)} & \textbf{0.817(0.045)} & 0.487(0.083) & 0.650(0.142) & 0.7285 \\
      P-IRM $(p=10)$ & 0.710(0.107) & 0.813(0.147) & 0.727(0.087) & \textbf{0.690(0.184)} & \textbf{0.735} \\
      % P-IRM $(p=25)$ & 0.820(0.148) & 0.742(0.138) & 0.597(0.243) & \textbf{0.753(0.209)} & 0.728 \\
      \midrule
      IB\_IRM & 0.647(0.197) & 0.740(0.171) & \textbf{0.750(0.155)} & 0.303(0.241) & 0.61 \\
      P-IB\_IRM $(p=0)$ & 0.663(0.242) & 0.643(0.137) & 0.437(0.289) & 0.617(0.059) & 0.59 \\
      P-IB\_IRM $(p=10)$ & \textbf{0.690(0.340)} & \textbf{0.790(0.070)} & 0.377(0.214) & \textbf{0.837(0.160)} & \textbf{0.6735} \\
      % P-IB\_IRM $(p=25)$ & 0.613(0.386) & 0.740(0.171) & 0.203(0.029) & 0.343(0.464) & 0.47475 \\
      \bottomrule
      \end{tabular}%
      \normalsize	
      % \vspace{-6pt}
     \caption{Domaing Generalization in Metashift. Training environments are $d$ away from testing community $dog(shelf)$, with additional samples up to percentage $p \in \{0, 10\}$ for partitioned models. Results for $p=25$ are in Table 5 (in Appendix)}.
     %Communities in training are not observed during testing. 
     %Results for $p=25$ are in Table 5\ref{tab:metashift_domain_generalization_DG_p_25} (in Appendix)}
    \label{tab:metashift_domain_generalization_DG}%
    % \vspace{-6pt}
  \end{table*}%

\textbf{Results} For all experiments, we report the binary classification accuracy averaged over 3 seeds, with the randomness solely arising from the learning algorithm. We compare the performance of P-IRM against IRM, as well other benchmarks and their corresponding partitioned versions in table~\ref{tab:metashift_domain_generalization_DG}. In most experiments, especially those with higher deviation between the training and testing data, models with partitioning tend to perform better. 

\subsection{Named Entity Recognition (NER)}
Distributional shifts are common in language tasks, given that societal changes are known to influence language usage over time. These changes are also reflected in word embeddings (words vectors to represent language) \cite{gargSJ2018_100yrswords}. Within this context, we explore effects of partitioning \cite{Lazaridou2021_language_DAPT,luu2021time_embeddings}. \\ \textbf{Experiment Setup} We consider the SciERC \cite{luan2018sciERCdataset} dataset, which consists of CS publications from 1980 to 2016. The specific task is \textit{Named Entity Recognition}, a multi-class classification task, that labels each scientific mention in a sentence into six possible categories (\textit{Task}, \textit{Method}, \textit{Evaluation Metric,  \textit{Material}, \textit{Other-Scientific-Term}, or \textit{Generic}}).The training set comprises of years from 1980-2009 and we test the model on data obtained between 2010-2016, with an intention to study distribution shift over time. For creating the training environments, we split training years into smaller intervals, 1990-2009, 2000-2009 and 2005-2009, such that each interval has roughly the same number of samples. For partitioning, we consider contiguous partitions of time intervals, based on the intuition that vocabularies in text have higher overlap when closer in time \cite{gururangan2020don}. For building the model, we train a classifier over the BERT pretrained language model \cite{devlin2019bert}. Due to high sample complexity, we also consider the conditioned P-IRM method that makes use of all training environments. \\ \textbf{Results} We report the classification accuracy, averaged over $3$ seeds in table ~\ref{tab:apx_language_results}. We find that both variants of P-IRM indeed improve performance over IRM. Additionally, we find that leveraging more training data using conditioned P-IRM leads to marginally better predictors, when compared against standard partitioning. Comparisons against IB\_IRM as well as ERM demonstrate that partitioning can improve efficacy of other learning algorithms as well.   % Table generated by Excel2LaTeX from sheet 'language_hal'
%   \begin{table}[!ht]
%     \centering
    
%       \begin{tabular}{c|c|c}
%       \toprule
%       \textcolor[rgb]{ .129,  .129,  .129}{Model} & Training Years & Testing (2010-2016) Accuracy \\
%       \midrule
%       \textcolor[rgb]{ .129,  .129,  .129}{ERM} & 1980-2009 & \textbf{0.813(0.006)} \\
%       IRM & 1980-2009 & \textcolor[rgb]{ .129,  .129,  .129}{0.800(0.008)} \\
%       P-IRM (partitioned) & 1990-2009 & \textcolor[rgb]{ .114,  .11,  .114}{0.803(0.014)} \\
%       P-IRM (partitioned) & 2000-2009 & \textcolor[rgb]{ .114,  .11,  .114}{0.805(0.007)} \\
%       P-IRM (conditioned) & 1990-2009 & \textcolor[rgb]{ .129,  .129,  .129}{0.801(0.014)} \\
%       P-IRM (conditioned) & 2000-2009 & \textcolor[rgb]{ .129,  .129,  .129}{0.808(0.012)} \\
%       \bottomrule
%       \end{tabular}%
%       \vspace{6pt}
%       \caption{Testing results (averaged over five seeds) on the NER dataset with batch size 2 shows that while P-IRM vs IRM trends remain similar, the IRM performance in general drops.}
%     \label{tab:apx_language_results}%
%   \end{table}%
  \begin{table}[!ht]
  \small
    \centering
    \tabcolsep=0.01cm
  % \scalebox{0.68}{
      \begin{tabular}{cccc}
      \toprule
      Model & \# envs & Training & Accuracy (2010-2016) \\
      \midrule
      ERM & 4   & 1980-2009 & 0.800 (0.012) \\
      P-ERM & 3   & 1990-2009 & \textbf{0.804 (0.020)} \\
      P-ERM & 2   & 2000-2009 & 0.804 (0.016) \\
      \midrule
      IRM & 4   & 1980-2009 & 0.795 (0.005) \\
      P-IRM (partitioned) & 3   & 1990-2009 & 0.795 (0.017) \\
      P-IRM (partitioned) & 2   & 2000-2009 & 0.807 (0.005) \\
      P-IRM (conditioned) & 3   & 1990-2009 & \textbf{0.812 (0.008)} \\
      P-IRM (conditioned) & 2   & 2000-2009 & 0.807 (0.015) \\
      \midrule
      IB\_IRM & 4   & 1980-2009 & 0.800 (0.010) \\
      P-IB\_IRM (partitioned) & 3   & 1990-2009 & 0.800 (0.015) \\
      P-IB\_IRM (partitioned) & 2   & 2000-2009 & 0.794 (0.015) \\
      P-IB\_IRM (conditioned) & 3   & 1990-2009 & \textbf{0.807 (0.008)} \\
      P-IB\_IRM (conditioned) & 2   & 2000-2009 & 0.805 (0.020) \\
      \bottomrule
      \end{tabular}%
      % }
      % \vspace{-6pt}
      \normalsize	
      \caption{Language Shifts in SciERC dataset. Partitioning improves performance, with (1990-2009) consistently optimal across all learning algorithms. 
      % A consistent optimal partition (1990-2009) across algorithms.
      % The partitioning improves performance for not only IRM but also other learning objectives, with the choice of optimal partition (1990-2009) consistent across training algorithms
      }
    \label{tab:apx_language_results}%
    % \vspace{-6pt}
  \end{table}%

 %\vspace{-1.6em}
\section{Discussion}\label{sec:discussion}
In this work, we propose partial invariance, as a relaxation of IRM objective, which allows us to explore a subtle trade-off in invariant models, namely accessing more domains at the cost of a smaller permissible invariant feature set. We then verify, with experiments across multiple domains, that when feasible, partitioning can indeed improve upon IRM as well as other learning frameworks. \\
We note that the proposed framework is naturally limited by the available information about training/deployment domains. While distribution shifts across time allows for partitions to be contiguous time intervals, finding appropriate partitions is non-trivial under complex shift topologies. In that sense, our work is the first step towards understanding the need for training domain selection in invariant learning. Thus, developing general heuristics for identifying the right partition is an important direction of future work. Second, we note that the conditional variant of P-IRM provides tangible gains in low data regimes, and it is of interest to study the nature of the accessible feature set as well as the associated sample complexities.

\bibliography{full,conf_full, pirm}

\begin{thebibliography}{52}
\providecommand{\natexlab}[1]{#1}

\bibitem[{Adragna et~al.(2020)Adragna, Creager, Madras, and
  Zemel}]{robertACMZ2020}
Adragna, R.; Creager, E.; Madras, D.; and Zemel, R.~S. 2020.
\newblock Fairness and Robustness in Invariant Learning: {A} Case Study in
  Toxicity Classification.
\newblock \emph{CoRR}, abs/2011.06485.

\bibitem[{Ahuja et~al.(2021)Ahuja, Caballero, Zhang, Gagnon-Audet, Bengio,
  Mitliagkas, and Rish}]{kartikACZGBMR2021}
Ahuja, K.; Caballero, E.; Zhang, D.; Gagnon-Audet, J.-C.; Bengio, Y.;
  Mitliagkas, I.; and Rish, I. 2021.
\newblock Invariance Principle Meets Information Bottleneck for
  Out-of-Distribution Generalization.
\newblock In \emph{Advances in Neural Information Processing Systems}.

\bibitem[{Ahuja et~al.(2020{\natexlab{a}})Ahuja, Shanmugam, Varshney, and
  Dhurandhar}]{ahujaSVD2020}
Ahuja, K.; Shanmugam, K.; Varshney, K.~R.; and Dhurandhar, A.
  2020{\natexlab{a}}.
\newblock Invariant {R}isk {M}inimization {G}ames.
\newblock In \emph{Proceedings of the 37th International Conference on Machine
  Learning (ICML'20)}, volume 119, 145--155. PMLR.

\bibitem[{Ahuja et~al.(2020{\natexlab{b}})Ahuja, Wang, Dhurandhar, Shanmugam,
  and Varshney}]{ahuja_erm_irm2020}
Ahuja, K.; Wang, J.; Dhurandhar, A.; Shanmugam, K.; and Varshney, K.~R.
  2020{\natexlab{b}}.
\newblock Empirical or {I}nvariant {R}isk {M}inimization? {A} {S}ample
  {C}omplexity {P}erspective.
\newblock In \emph{Proceeding of the 8th International Conference on Learning
  Representations (ICLR'20)}.

\bibitem[{Arjovsky et~al.(2019)Arjovsky, Bottou, Gulrajani, and
  Lopez-Paz}]{arjovskyBGP2020}
Arjovsky, M.; Bottou, L.; Gulrajani, I.; and Lopez-Paz, D. 2019.
\newblock Invariant {R}isk {M}inimization.
\newblock arXiv:1907.02893 [stat.ML].

\bibitem[{Beery, V.~Horn, and Perona(2018)}]{beeryGP2018}
Beery, S.; V.~Horn, G.; and Perona, P. 2018.
\newblock Recognition in {T}erra {I}ncognita.
\newblock In \emph{Proceedings of the European Conference on Computer Vision
  (ECCV)}, 456--473.

\bibitem[{Belkin and Niyogi(2001)}]{spectralEmbeddings}
Belkin, M.; and Niyogi, P. 2001.
\newblock Laplacian eigenmaps and spectral techniques for embedding and
  clustering.
\newblock \emph{Advances in neural information processing systems}, 14.

\bibitem[{Bellot and van~der Schaar(2020)}]{alexisBV2020}
Bellot, A.; and van~der Schaar, M. 2020.
\newblock Accounting for Unobserved Confounding in Domain Generalization.
\newblock arXiv:2007.10653 [stat.ML].

\bibitem[{Ben-David et~al.(2010)Ben-David, Blitzer, Crammer, Kulesza, Pereira,
  and Vaughan}]{shaiBCKPV2010}
Ben-David, S.; Blitzer, J.; Crammer, K.; Kulesza, A.; Pereira, F.; and Vaughan,
  J. 2010.
\newblock A theory of learning from different domains.
\newblock \emph{Machine Learning}, 79: 151--175.

\bibitem[{Ben-Tal, El~Ghaoui, and Nemirovski(2009)}]{benTEN2009}
Ben-Tal, A.; El~Ghaoui, L.; and Nemirovski, A. 2009.
\newblock \emph{Robust Optimization}.
\newblock Princeton Series in Applied Mathematics. Princeton University Press.

\bibitem[{Devlin et~al.(2019)Devlin, Chang, Lee, and
  Toutanova}]{devlin2019bert}
Devlin, J.; Chang, M.-W.; Lee, K.; and Toutanova, K. 2019.
\newblock BERT: Pre-training of Deep Bidirectional Transformers for Language
  Understanding.
\newblock In \emph{Proceedings of the 2019 Conference of the North American
  Chapter of the Association for Computational Linguistics: Human Language
  Technologies, Volume 1 (Long and Short Papers)}, 4171--4186.

\bibitem[{Duchi, Glynn, and Namkoong(2021)}]{duchiGN2018}
Duchi, J.; Glynn, P.; and Namkoong, H. 2021.
\newblock Statistics of {Robust} {Optimization}: {A} {Generalized} {Empirical}
  {Likelihood} {Approach}.
\newblock \emph{Mathematics of Operations Research}, 46(3).

\bibitem[{Gallacher(2021)}]{gallacher2021}
Gallacher, J.~D. 2021.
\newblock Leveraging cross-platform data to improve automated hate speech
  detection.
\newblock arXiv:2102.04895 [CS.CL].

\bibitem[{Garg et~al.(2018)Garg, Schiebinger, Jurafsky, and
  Zou}]{gargSJ2018_100yrswords}
Garg, N.; Schiebinger, L.; Jurafsky, D.; and Zou, J. 2018.
\newblock Word embeddings quantify 100 years of gender and ethnic stereotypes.
\newblock \emph{Proceedings of the National Academy of Sciences}, 115(16):
  E3635--E3644.

\bibitem[{Gong et~al.(2016)Gong, Zhang, Liu, Tao, Glymour, and
  Sch\"{o}lkopf}]{gongZLTGS2016}
Gong, M.; Zhang, K.; Liu, T.; Tao, D.; Glymour, C.; and Sch\"{o}lkopf, B. 2016.
\newblock Domain {A}daptation with {C}onditional {T}ransferable {C}omponents.
\newblock In \emph{Proceedings of the 33rd International Conference on Machine
  Learning (ICML'16)}, volume~48, 2839--2848.

\bibitem[{Gulrajani and Lopez-Paz(2020)}]{gulrajaniL2020}
Gulrajani, I.; and Lopez-Paz, D. 2020.
\newblock In {S}earch of {L}ost {D}omain {G}eneralization.
\newblock In \emph{Proceeding of the 8th International Conference on Learning
  Representations (ICLR'20)}.

\bibitem[{Gulrajani and Lopez{-}Paz(2020)}]{ishaanGP2020}
Gulrajani, I.; and Lopez{-}Paz, D. 2020.
\newblock In Search of Lost Domain Generalization.
\newblock \emph{CoRR}, abs/2007.01434.

\bibitem[{Gururangan et~al.(2020)Gururangan, Marasovi{\'c}, Swayamdipta, Lo,
  Beltagy, Downey, and Smith}]{gururangan2020don}
Gururangan, S.; Marasovi{\'c}, A.; Swayamdipta, S.; Lo, K.; Beltagy, I.;
  Downey, D.; and Smith, N.~A. 2020.
\newblock Don’t Stop Pretraining: Adapt Language Models to Domains and Tasks.
\newblock In \emph{Proceedings of the 58th Annual Meeting of the Association
  for Computational Linguistics}, 8342--8360.

\bibitem[{He et~al.(2016)He, Zhang, Ren, and Sun}]{he2016resnet}
He, K.; Zhang, X.; Ren, S.; and Sun, J. 2016.
\newblock Deep residual learning for image recognition.
\newblock In \emph{Proceedings of the IEEE conference on computer vision and
  pattern recognition}, 770--778.

\bibitem[{Heinze-Deml, Peters, and Meinshausen(2018)}]{heinzeCM2018}
Heinze-Deml, C.; Peters, J.; and Meinshausen, N. 2018.
\newblock Invariant {C}ausal {P}rediction for {N}onlinear {M}odels.
\newblock \emph{Journal of Causal Inference}, 6(2).

\bibitem[{Jin, Barzilay, and Jaakkola(2020)}]{wengongJBJ2020}
Jin, W.; Barzilay, R.; and Jaakkola, T.~S. 2020.
\newblock Domain Extrapolation via Regret Minimization.
\newblock \emph{CoRR}, abs/2006.03908.

\bibitem[{Kamath~Pritish and Srebro(2021)}]{KamathTDS2021}
Kamath~Pritish, D.~S., Akilesh~Tangella; and Srebro, N. 2021.
\newblock Does {I}nvariant {R}isk {M}inimization {C}apture {I}nvariance?
\newblock In \emph{Proceedigns of the International Conference on Artificial
  Intelligence and Statistics}, 4069--4077. PMLR.

\bibitem[{Kingma and Ba(2015)}]{diederikKB2015}
Kingma, D.~P.; and Ba, J. 2015.
\newblock Adam: {A} Method for Stochastic Optimization.
\newblock In Bengio, Y.; and LeCun, Y., eds., \emph{3rd International
  Conference on Learning Representations, {ICLR} 2015, San Diego, CA, USA, May
  7-9, 2015, Conference Track Proceedings}.

\bibitem[{Koh et~al.(2021)Koh, Sagawa, Marklund, Xie, Zhang, Balsubramani, Hu,
  Yasunaga, Phillips, Gao, Lee, David, Stavness, Guo, Earnshaw, Haque, Beery,
  Leskovec, Kundaje, Pierson, Levine, Finn, and Liang}]{koh_wilds2019}
Koh, P.~W.; Sagawa, S.; Marklund, H.; Xie, S.~M.; Zhang, M.; Balsubramani, A.;
  Hu, W.; Yasunaga, M.; Phillips, R.~L.; Gao, I.; Lee, T.; David, E.; Stavness,
  I.; Guo, W.; Earnshaw, B.; Haque, I.; Beery, S.~M.; Leskovec, J.; Kundaje,
  A.; Pierson, E.; Levine, S.; Finn, C.; and Liang, P. 2021.
\newblock WILDS: A Benchmark of in-the-Wild Distribution Shifts.
\newblock In Meila, M.; and Zhang, T., eds., \emph{Proceedings of the 38th
  International Conference on Machine Learning}, volume 139 of
  \emph{Proceedings of Machine Learning Research}, 5637--5664. PMLR.

\bibitem[{Krueger et~al.(2021)Krueger, Caballero, Jacobsen, Zhang, Binas,
  Zhang, Le~Priol, and Courville}]{kruegerCJ2021}
Krueger, D.; Caballero, E.; Jacobsen, J.-H.; Zhang, A.; Binas, J.; Zhang, D.;
  Le~Priol, R.; and Courville, A. 2021.
\newblock Out-of-{D}istribution {G}eneralization via {R}isk {E}xtrapolation.
\newblock In \emph{Proceedings of the 38th International Conference on Machine
  Learning (ICML'21)}, 5815--5826. PMLR.

\bibitem[{Lake et~al.(2017)Lake, Ullman, Tenenbaum, and
  Gershman}]{lakeRTGS2017}
Lake, B.~M.; Ullman, T.~D.; Tenenbaum, J.~B.; and Gershman, S.~J. 2017.
\newblock Building machines that learn and think like people.
\newblock \emph{Behavioral and Brain Sciences}, 40: e253.

\bibitem[{Lazaridou et~al.(2021)Lazaridou, Kuncoro, Gribovskaya, Agrawal,
  Liska, Terzi, Gimenez, de~Masson~d'Autume, Ruder, Yogatama, Cao, Kociský,
  Young, and Blunsom}]{Lazaridou2021_language_DAPT}
Lazaridou, A.; Kuncoro, A.; Gribovskaya, E.; Agrawal, D.; Liska, A.; Terzi, T.;
  Gimenez, M.; de~Masson~d'Autume, C.; Ruder, S.; Yogatama, D.; Cao, K.;
  Kociský, T.; Young, S.; and Blunsom, P. 2021.
\newblock Pitfalls of Static Language Modelling.
\newblock arXiv:2102.01951 [cs.CL].

\bibitem[{Lee and Raginsky(2018)}]{leeR2018}
Lee, J.; and Raginsky, M. 2018.
\newblock Minimax {Statistical} {Learning} with {Wasserstein} {Distances}.
\newblock In \emph{Proceedings of the 32nd International Conference on Neural
  Information Processing Systems (NIPS'18)}, 2692--2701.

\bibitem[{Li et~al.(2018)Li, Gong, Tian, Liu, and Tao}]{liGTLT2018}
Li, Y.; Gong, M.; Tian, X.; Liu, T.; and Tao, D. 2018.
\newblock Domain {G}eneralization via {C}onditional {I}nvariant
  {R}epresentation.
\newblock \emph{Proceedings of the 32nd Association for the Advancement of
  Artificial Intelligence (AAAI'18)}, 31(1).

\bibitem[{Liang and Zou(2022)}]{liang2022metashift}
Liang, W.; and Zou, J. 2022.
\newblock Metashift: A dataset of datasets for evaluating contextual
  distribution shifts and training conflicts.
\newblock In \emph{International Conference on Learning Representations, {ICLR}
  2022}.

\bibitem[{Lin, Zhu, and Cui(2022)}]{lin2022zin}
Lin, Y.; Zhu, S.; and Cui, P. 2022.
\newblock ZIN: When and How to Learn Invariance by Environment Inference?
\newblock \emph{arXiv preprint arXiv:2203.05818}.

\bibitem[{Luan et~al.(2018)Luan, He, Ostendorf, and
  Hajishirzi}]{luan2018sciERCdataset}
Luan, Y.; He, L.; Ostendorf, M.; and Hajishirzi, H. 2018.
\newblock Multi-Task Identification of Entities, Relations, and Coreferencefor
  Scientific Knowledge Graph Construction.
\newblock In \emph{Proc.\ Conf. Empirical Methods Natural Language Process.
  (EMNLP)}.

\bibitem[{Luu et~al.(2021)Luu, Khashabi, Gururangan, Mandyam, and
  Smith}]{luu2021time_embeddings}
Luu, K.; Khashabi, D.; Gururangan, S.; Mandyam, K.; and Smith, N.~A. 2021.
\newblock Time Waits for No One! Analysis and Challenges of Temporal
  Misalignment.
\newblock ArXiv preprint arXiv:2111.07408.

\bibitem[{Mahajan, Tople, and Sharma(2021)}]{mahajanTS2021}
Mahajan, D.; Tople, S.; and Sharma, A. 2021.
\newblock Domain {Generalization} using {Causal} {Matching}.
\newblock In \emph{Proceedings of the 38th International Conference on Machine
  Learning (ICML'21)}, volume 139, 7313--7324. PMLR.

\bibitem[{Mani, Varshney, and Pentland(2021)}]{ManiV2021}
Mani, A.; Varshney, L.~R.; and Pentland, A. 2021.
\newblock Quantization {G}ames on {S}ocial {N}etworks and {L}anguage
  {E}volution.
\newblock \emph{IEEE Transactions on Signal Processing}, 69: 3922--3934.

\bibitem[{Marcus(2018)}]{marcus2018}
Marcus, G. 2018.
\newblock Deep {L}earning: {A} {C}ritical {A}ppraisal.
\newblock arXiv:1801.00631 [CS.AI].

\bibitem[{Mohri, Sivek, and Suresh(2019)}]{mohriSS2019}
Mohri, M.; Sivek, G.; and Suresh, A.~T. 2019.
\newblock Agnostic {Federated} {Learning}.
\newblock In \emph{Proceedings of the 36th International Conference on Machine
  Learning (ICML'19)}, volume~97, 4615--4625. PMLR.

\bibitem[{Neal(1996)}]{nealR1996}
Neal, R.~M. 1996.
\newblock \emph{Bayesian Learning for Neural Networks}.
\newblock Berlin, Heidelberg: Springer-Verlag.
\newblock ISBN 0387947248.

\bibitem[{Ng, Jordan, and Weiss(2001)}]{ng2001spectral}
Ng, A.; Jordan, M.; and Weiss, Y. 2001.
\newblock On spectral clustering: Analysis and an Algorithm.
\newblock \emph{Advances in Neural Information Processing Systems}, 14.

\bibitem[{Pearl(2009)}]{judeaP2009}
Pearl, J. 2009.
\newblock {Causal inference in statistics: An overview}.
\newblock \emph{Statistics Surveys}, 3(none): 96 -- 146.

\bibitem[{Peters, B{\"u}hlmann, and Meinshausen(2016)}]{petersBM2016}
Peters, J.; B{\"u}hlmann, P.; and Meinshausen, N. 2016.
\newblock Causal inference by using invariant prediction: identification and
  confidence intervals.
\newblock \emph{Journal of the Royal Statistical Society. Series B (Statistical
  Methodology)}, 78(5): 947--1012.

\bibitem[{Peters, Bühlmann, and Meinshausen(2015)}]{jonasPBM2015}
Peters, J.; Bühlmann, P.; and Meinshausen, N. 2015.
\newblock Causal inference using invariant prediction: identification and
  confidence intervals.
\newblock Preprint.

\bibitem[{Peyrard et~al.(2021)Peyrard, Ghotra, Josifoski, Agarwal, Patra,
  Carignan, Kiciman, and West}]{maximePGJAPCKW2021}
Peyrard, M.; Ghotra, S.~S.; Josifoski, M.; Agarwal, V.; Patra, B.; Carignan,
  D.; Kiciman, E.; and West, R. 2021.
\newblock Invariant Language Modeling.
\newblock \emph{CoRR}, abs/2110.08413.

\bibitem[{Radford et~al.(2019)Radford, Wu, Child, Luan, Amodei, Sutskever
  et~al.}]{radford2019language}
Radford, A.; Wu, J.; Child, R.; Luan, D.; Amodei, D.; Sutskever, I.; et~al.
  2019.
\newblock Language models are unsupervised multitask learners.
\newblock \emph{OpenAI blog}, 1(8): 9.

\bibitem[{Rosenfeld, Ravikumar, and Risteski(2020)}]{rosenfeldRP2020}
Rosenfeld, E.; Ravikumar, P.~K.; and Risteski, A. 2020.
\newblock The {R}isks of {I}nvariant {R}isk {M}inimization.
\newblock In \emph{Proceeding of the 8th International Conference on Learning
  Representations (ICLR'20)}.

\bibitem[{Sanh et~al.(2019)Sanh, Debut, Chaumond, and
  Wolf}]{sanh2019distilbert}
Sanh, V.; Debut, L.; Chaumond, J.; and Wolf, T. 2019.
\newblock DistilBERT, a distilled version of BERT: smaller, faster, cheaper and
  lighter.

\bibitem[{Sch{\"o}lkopf et~al.(2012)Sch{\"o}lkopf, Janzing, Peters, Sgouritsa,
  Zhang, and Mooij}]{scholkop2012}
Sch{\"o}lkopf, B.; Janzing, D.; Peters, J.; Sgouritsa, E.; Zhang, K.; and
  Mooij, J. 2012.
\newblock On causal and {A}nticausal {L}earning.
\newblock In \emph{Proceedings of the 29th International Conference on Machine
  Learning (ICML'12)}, 1255--1262.

\bibitem[{Shui, Wang, and Gagn{\'{e}}(2021)}]{ChangjianSWG2021}
Shui, C.; Wang, B.; and Gagn{\'{e}}, C. 2021.
\newblock On the benefits of representation regularization in invariance based
  domain generalization.
\newblock \emph{CoRR}, abs/2105.14529.

\bibitem[{Vapnik(2013)}]{vapnik2013}
Vapnik, V. 2013.
\newblock \emph{The {N}ature of {S}tatistical {L}earning {T}heory}.
\newblock Springer Science and Business Media.

\bibitem[{Xie et~al.(2021)Xie, Ye, Chen, Liu, Sun, and Li}]{xieYCLSL2021}
Xie, C.; Ye, H.; Chen, F.; Liu, Y.; Sun, R.; and Li, Z. 2021.
\newblock Risk {Variance} {Penalization}.
\newblock arXiv:2006.07544 [cs.LG].

\bibitem[{Yu et~al.(2022)Yu, Zhu, Li, Hong, Zhang, Ye, Huang, and
  He}]{RunpengYZLHZYHH2022}
Yu, R.; Zhu, H.; Li, K.; Hong, L.; Zhang, R.; Ye, N.; Huang, S.-L.; and He, X.
  2022.
\newblock Regularization Penalty Optimization for Addressing Data Quality
  Variance in OoD Algorithms.
\newblock \emph{Proceedings of the AAAI Conference on Artificial Intelligence},
  36(8): 8945--8953.

\bibitem[{Zhang, Gong, and Schoelkopf(2015)}]{kunGS2015}
Zhang, K.; Gong, M.; and Schoelkopf, B. 2015.
\newblock Multi-{S}ource {D}omain {A}daptation: {A} {C}ausal {V}iew.
\newblock In \emph{Proceedings of the 29th Association for the Advancement of
  Artificial Intelligence Conference (AAAI'15)}.

\end{thebibliography}

\clearpage

\appendix

\section{Proofs}
\label{sec:appendix}

\subsection{Proof of Lemma 1}
\label{apx:lemma1proof}
The proof follows as a consequence of the parametrization. Under the MSE loss, note that the expected risk takes the following form: 
\begin{align}
& {R}^e(w \circ \Phi) = \expect_{\bm{y}^e, \bm{X}^e}(\bm{y}^e - w.\Phi^\top \bm{X}^e)^2 \mbox{,}
\end{align}
wherein the scalar $w$ is fixed at 1.0. The gradient penalty w.r.t. $w$ in environment $e$ can then be obtained as:
\begin{align}
\begin{split}
&\ \|\nabla_{w, w=1.0}{R}^{e'}(w \circ \Phi)\|= |\nabla_{w, w=1.0} \expect_{\bm{y}^e, \bm{X}^e}(\bm{y}^e - w.\Phi^\top \bm{X}^e)^2 | \\
&= |\expect_{\bm{y}^e, \bm{X}^e}[\nabla_{w, w=1.0}(\bm{y}^e - w.\Phi^\top \bm{X}^e)]^2| \\
&= |\expect_{\bm{y}^e, \bm{X}^e}[2(w.\Phi^\top \bm{X}^e - \bm{y}^e)\Phi^\top \bm{X}^e]| \\
&= |\expect_{{\epsilon_y}^e, \bm{X}^e}[2(\Phi - W^e)^\top \bm{X}^e - {\epsilon_y}^e){\bm{X}^e}^\top \Phi]|\\
&= |\expect_{\bm{X}^e}[2(\Phi - W^e)^\top (\bm{X}^e{\bm{X}^e}^\top) \Phi] + 0|\\
&= |2(\Phi - W^e)^\top \expect_{\bm{X}^e}[(\bm{X}^e{\bm{X}^e}^\top)] \Phi|\\
&= |2(\Phi - W^e)^\top \Phi|\\
&= 2|\sum_{i=1}^c(\Phi_i^2 - w^e_i.\Phi_i)|\mbox{,}\\
\end{split}
\end{align}
wherein the simplifications follow through due to feature independence, normalization and zero mean noise assumption. 
Notice that as per the constraints in \eqref{eq_irmgc}, we need the risk penalty term above to be equal to zero for all training environments:
$$|\sum_{i=1}^c(\Phi_i^2 - w^e_i.\Phi_i)| = 0\ \forall\ e \in \mathcal{E}_{tr}.$$ 
Then note from a risk minimization incentive, we naturally have $\Phi_1 = w_{inv}$ without incurring any penalty. Thus the penalty boils to:
$$|\sum_{i=2}^c(\Phi_i^2 - w^e_i.\Phi_i)| = 0\ \forall\ e \in \mathcal{E}_{tr}.$$
But from Assumption \ref{assump:irm}, we have an environment $e$ in the training set in which IRM is sufficient, i.e. $w^e_i = 0\ \forall\ i\neq 1$. Thus, the constraint in that environment equates to:
$$|\sum_{i=2}^c(\Phi_i^2 - 0.\Phi_i)| = |\sum_{i=2}^c \Phi_i^2| = 0\ \forall\ e \in \mathcal{E}_{tr}.$$
Thus to satisfy this constraint, $\Phi_i = 0 \ \forall\ i \neq 1$, which means the set of feasible solutions is comprised solely of the perfectly invariant predictor, as claimed.

\subsection{Proof of Lemma 2}

Under the uniform feature model, we sample $\mathcal{E}_{tr}$ such that $|\mathcal{E}_{tr}| = t$. Notice that the cardinality $m$ of the required partition $\mathcal{E}_{partition} := \{e \in \mathcal{E}_{tr}| \omega(e) = 1\}$ is random variable. \\
We note that for an environment sample $e \in \mathcal{E}_{tr}$, for a given reference environment $e^{ref}$, we have from our analysis in %\nameref{apx:them1}
the previous theorem that $P(\|W^e - W^{e_{ref}}\| \leq \delta) = P(E_1) + P(E_2)$ and that $P(E_1) > p P(E_2)$, wherein $p \geq \frac{(c-1-\delta)\alpha}{\delta} >> 1$. Let $n = c-2$ for brevity. Then, we have the following approximation: \begin{align*}&P(\|W^e - W^{e_{ref}}\| \leq \delta) \approx  P(E_1) \\ &\geq \gamma = \frac{1}{k\sqrt{2n}}\exp(-n D(\delta/n \| 1/\alpha k)),\end{align*}  
where the final result follows from standard anti-concentration bounds on a Binomial distribution. Then it is easy to see that given $|\mathcal{E}_{tr}| = t$, $P(|\mathcal{E}_{partition}| \geq m) = P(|\sum_{e \in \mathcal{E}_{tr}} \bm{1}[\omega(e) = 1] \geq m)] \geq P(\sum_{j=1}^t D_j \geq m]$ where $D_j$ is Bernoulli random variable distributed as $D_j \sim \text{Bern}(\gamma)$.

\noindent On the RHS, we get a sum of i.i.d. Bernoulli variables for and is thus, $\sum_{j=1}^t D_j \sim \text{Bin}(n, \gamma)$. Let $M = \sum_{j=1}^t D_j$ and notice that we need that $M > m$ with high probability. To achieve this, we first upper bound the probability of event $M < m$ using Chernoff's tail bound and derive conditions under which this upper bound is small. Specifically:

\begin{equation}
\label{eq:chernoff}
    P(M < m) < \text{exp}(-tD\bigg(\frac{m}{t} \| \gamma\bigg)) < \epsilon, 
\end{equation}
where $D(a\|b) = a\log(\frac{a}{b}) + (1-a)\log(\frac{1-a}{1-b})$. Next, assume that $t = Cm$. Then we can simplify the inequality:
\begin{subequations}
\begin{align}
-Cm\bigg(\frac{1}{C}\log\frac{\frac{1}{\gamma}}{C} + \frac{C-1}{C}\log\frac{(C-1) \frac{1}{\gamma}}{C( \frac{1}{\gamma}-1)}\bigg) < \log(\epsilon)\\
=> \bigg(\log\frac{ \frac{1}{\gamma}}{C} + ({C-1})\log\frac{(C-1) \frac{1}{\gamma}}{C( \frac{1}{\gamma}-1)}\bigg) > 1/m\log(1/\epsilon)\\
=> \bigg( C\log\frac{(C-1) \frac{1}{\gamma}}{C( \frac{1}{\gamma}-1)} - \log\frac{C-1}{ \frac{1}{\gamma}-1}\bigg) > 1/m\log(1/\epsilon)\\
\asymp \bigg( C\log\frac{(C-1) \frac{1}{\gamma}}{C( \frac{1}{\gamma}-1)}\bigg) > 1/m\log(1/\epsilon).
\end{align}
\end{subequations}

\noindent Our inequality will be satisfied if a) $C >  \frac{1}{\gamma}\log(1/\epsilon)$ and b) $\log(\frac{(C-1) \frac{1}{\gamma}}{C( \frac{1}{\gamma}-1)}) > \gamma m$. Then we can show that for sufficiently large $ \frac{1}{\gamma} m$, condition b) roughly amounts $C > c_i(1+\frac{1}{m-1})$.  So if $C > \max\{ \frac{1}{\gamma}\log(1/\epsilon),  \frac{1}{\gamma}(1+\frac{1}{m-1}\} \sim  \frac{1}{\gamma}\log(1/\epsilon)$ (for small $\epsilon$), then we have $P(M < m) < \epsilon$. Hence, for $t = Cm \sim  \frac{1}{\gamma} m \log(1/\epsilon),$ we get that $M = \sum_{j=1}^t D_j$ with high probability. But note that we  already have $P(|\mathcal{E}_{partition}| \geq m) \geq P(\sum_{j=1}^t D_j \geq m]$. Thus, $ P(|\mathcal{E}_{partition}| \geq m) > 1 - \epsilon$.

\clearpage

\subsection{Proof of Theorem 1}
\label{apx:them1}

 We begin by restating the theorem.\\
  \begin{theorem*}$\bm{1}$ 
 \label{thmproof:1}
     %Under the assumed model \eqref{eq:gen_model}, along with access to oracle $\omega(e)$ with $\delta < (c-2)/2$, isolate $\mathcal{E}_{partition} := \{e \in \mathcal{E}_{tr}| \omega(e) = 1\} \cup \{e_{ref}\} \subseteq \mathcal{E}_{tr}$. Next, let $|A_i| = k$, where $A_i$ is the set corresponding to the feature weight $w_i^{e_{ref}}$ of interest. Then, if the sets $\{A_j\} \forall\ j \in \{2, \ldots, c\} \setminus \ \{i\}$ satisfy $|A_j| > \alpha k$ for some $\alpha > 1$ and as before, in the new set of training environments  $\mathcal{E}_{partition}$, $\exists (e, e')\in \mathcal{E}_{partition}$ such that for the weight for each feature $j \in \{2, \ldots, c\}\setminus \ \{i\}$, satisfies ${w^e}_i < {w^{e'}}_i$, %satisfies that $|A_{j}^{partition}| \geq 2\  \forall\ j \in \{2, \ldots, c\} \setminus \ \{i\}$, 
     %we have with probability greater or equal to $(\frac{p}{p+1})^{|\mathcal{E}_{partition}|}$, where $p \geq \frac{(c-1 - \delta)\alpha}{\delta}$, the IRM solution over set $\mathcal{E}_{partition}$ will recover the feature of interest $w_i^{e_{ref}}$.  
     Under the model \eqref{eq:gen_model}, under Assumption \ref{assump:pirm}, with access to oracle $\omega(e) = \mathbf{1}[ \|W^{e_{ref}} - W^{e}\|_0 \leq \delta]$ and $\delta < (c-2)/2$, isolate $\mathcal{E}_{partition} := \{e \in \mathcal{E}_{tr}| \omega(e) = 1\} \cup \{e_{ref}\} \subseteq \mathcal{E}_{tr}$. Next, let $|A_i| = k$, where $A_i$ is the set corresponding to the feature weight $w_i^{e_{ref}}$ of interest. Then, if the sets $\{A_j\} \forall\ j \in \{2, \ldots, c\} \setminus \ \{i\}$ satisfy $|A_j| > \alpha k$ for some $\alpha > 1$, we have with probability greater or equal to $(\frac{p}{p+1})^{|\mathcal{E}_{partition}|}$, where $p \geq \frac{(c-1 - \delta)\alpha}{\delta}$, the IRM solution over set $\mathcal{E}_{partition}$ will recover the feature of interest $w_i^{e_{ref}}$.
    \end{theorem*}

\noindent The proof sketch is as follows. We first characterize the probability of error in using the oracle as an indicator for the partition membership. Assuming the partition is identified, we can then directly apply Lemma \ref{lemma:1} to obtain the desired result. We begin by characterizing the two possible cases that arise when oracle $\omega(e)$ = 1.  

\begin{itemize}
    \item $E_1$: If feature at level $i$ remains unchanged, this means that out of $c-2$ features (discarding the feature at level $i$ and the invariant feature at level $1$), a maximum of up-to $\delta$ features changed in the worst case. 
    \item $E_2$: If feature at level $i$ changed, this means that out of $c-2$ features (discarding the feature at level $i$ and the invariant feature at level $1$), a maximum of up-to $\delta-1$ features changed in the worst case. 
\end{itemize}

\noindent Note that both of the events can be modelled as a sum of Bernoulli random variables with different probabilities of success. Consider $B_j \sim \text{Bern}(1- 1/|A_j|)$, which indicates whether the feature value at level $j$ changed under the uniform model. In the first case, we model the conditional probability as:
\begin{align}
    \begin{split}
        &\ P(E_1) = P(w^{e}_i = w_i^{e_{ref}} |\omega(e)=1)\\ 
        &= (1/k) P(1 \leq \sum_{{j=2}, j \neq i}^c B_j \leq \delta),
    \end{split}
\end{align}
while in the alternate case,
\begin{align}
    \begin{split}
        &P(E_2) = P(w^{e}_i \neq  w_i^{e_{ref}}|\omega(e)=1) \\ 
        &= (1-1/k) P(0 \leq \sum_{{j=2}, j \neq i}^c B_j \leq \delta-1).
    \end{split}
\end{align}

\noindent Next, we make use of the assumption on cardinalities and analyze the probabilities corresponding to each term. We note that:
$$P(E_1) = \sum_{m=1}^\delta (1/k) P(\sum_{{j=2}, j \neq i}^c B_j = m).$$
Similarly:
$$P(E_2) = \sum_{m=0}^{\delta-1} (1-1/k) P(\sum_{{j=2}, j \neq i}^c B_j = m)$$

\noindent For brevity, we represent each individual term on the right hand side of the summations as $P_{a,m}$ $(a=1 \text{ or } 2 \text{ depending on the event})$, corresponding to that value of $m$. Next, we analyze the ratio for the final two terms in either sequence. Let $c-2 = n$ and note that using the assumption on cardinalities of the feature sets, we can bound the ratio as follows:
$$\frac{P_{1, \delta}}{P_{2, {\delta-1}}} \geq \frac{(1/k)({\binom{n}{\delta}}(1-1/\alpha k)^\delta (1/\alpha k)^{n-\delta})}{(1-1/k)({\binom{n}{\delta-1}}(1-1/\alpha k)^{\delta-1} (1/\alpha k)^{n+1-\delta})},$$
wherein the greater than equal to sign holds since cardinality of each feature set is greater than or equal to $\alpha k$. Simplifying the terms, we get:
$$\frac{P_{1, \delta}}{P_{2, {\delta-1}}} \geq \frac{({n+1- \delta)}(\alpha k-1) }{(k-1)({\delta)}} \geq \frac{(c-1 - \delta)\alpha}{\delta}= p.$$

\noindent Note that this ratio increases as $\delta$ reduces and therefore, we conclude that: $$P(E_1)/P(E_2) = \frac{\sum_{m=1}^\delta P_{1, m}}{\sum_{m=1}^{\delta-1} P_{2, m}} \geq p.$$

\noindent Thus, we obtain that $P_{error} = P(E_2) \leq 1/(p+1)$. Thus, for our training partition $\mathcal{E}_{partition} = \{e \in \mathcal{E}_{tr}|\omega(e) = 1\}$, with probability greater than $(\frac{p}{p+1})^{|\mathcal{E}_{partition}|}$, we will learn an accurate partition. Since within this partition the feature weight corresponding to $\bm{x}_i$ is same as the reference for all environments, we obtain the required result from Lemma \ref{lemma:1}.
 
 \clearpage
 
 \subsection{Proof of Theorem 2}

\label{thmproof:2}
First, we restate the theorem more formally.\\
\begin{theorem*}$\bm{2}$ (Formal)
 %Assume we observe $(\tilde{\bm{x}}^e, y^e)$ as per \eqref{eq:observe_model}, with environments $e \in \mathcal{E}_{tr}$ sampled as per \eqref{eq:gen_model} and let $\mathcal{E}_{partition} := \{e \in \mathcal{E}_{tr}| \omega(e) = 1\} \cup e_{ref} \subseteq$. Let $\bm{\Phi} \in \mathbb{R}^{d \times d}$ have rank $r > 0$. Then sampling $|\mathcal{E}_{tr}| >  \frac{1}{\gamma}(d-r+d/r)\log(1/\epsilon)$ ensures partition cardinality $|\mathcal{E}_{partition}| > d - r + d/r$ with probability > $1-\epsilon$. Furthermore, if $e \in \mathcal{E}_{partition}$ lie in linear general position of degree $r$ (Assumption 3 in Appendix), then with probability greater than or equal to $(\frac{p}{p+1})^{|\mathcal{E}_{partition}|}$, where $p \geq \frac{(c-1 - \delta)\alpha}{\delta}$, the oracle identifies $\mathcal{E}_{partition}$ such that the learnt $\bm{w} \circ \bm{\Phi}$ via IRM within the partition recovers the desired features/weights $(\bar{\bm{x}}_{inv}^e)^\top \bar{\bm{w}}_{inv}^e$,  $\forall e \in \mathcal{E}_{all}$ which satisfy $w^{e}_i = w^{e_{ref}}_i$
Assume we observe samples $(\tilde{\bm{X}}^e, \bm{y}^e)$ as per equation \eqref{eq:observe_model}, with environments $e \in \mathcal{E}_{tr}$ sampled as per equation \eqref{eq:gen_model} and let $\mathcal{E}_{partition} := \{e \in \mathcal{E}_{tr}| \omega(e) = 1\}$. Let $\bm{\Phi} \in \mathbb{R}^{d \times d}$ have rank $r$. Then sampling $|\mathcal{E}_{tr}| >  \frac{1}{\gamma}(d-r+d/r)\log(1/\epsilon)$ ensures that partition cardinality $|\mathcal{E}_{partition}| > d - r + d/r$ with probability > $1-\epsilon$. Furthermore, if $e \in \mathcal{E}_{tr}$ lie in linear general position of degree $r$ (Assumption 3), then with probability greater than or equal to $(\frac{p}{p+1})^{|\mathcal{E}_{partition}|}$, where $p \geq \frac{(c-1 - \delta)\alpha}{\delta}$, the oracle identifies $\mathcal{E}_{partition}$ such that we have:
$$\Phi \expect_{\tilde{\bm{X}}^e}[\tilde{\bm{X}}^e (\tilde{\bm{X}}^e)^\top] \Phi^\top W = \Phi \expect_{\tilde{\bm{X}}^e, \bm{y}^e}[\tilde{\bm{X}}^e \bm{y}^e],$$
holds for all $e \in \mathcal{E}_{partition}$ iff $\bm{\Phi}$ elicits an invariant predictor $\bm{\Phi}^\top W$ $\forall\ e \in \mathcal{E}_{all}$ whose feature weights satisfy $w^{e}_i = w^{e_{ref}}_i$. 
 \end{theorem*}

\noindent We begin by recollecting the requisite tools from \cite{arjovskyBGP2020}. \\ 
\begin{assumption}
\label{assump:3}
With observation model as in equation \eqref{eq:observe_model}, a set of training environments $\mathcal{E}_{partition} \subseteq \mathcal{E}_{tr}$ lie in linear general position of degree $r$ if $|\mathcal{E}_{partition}| > d-r+d/r$ for some $r \in \mathbb{N}$, $r<d$, and for all non-zero $X \in \mathbb{R}^d$:
\begin{align*}
&\text{dim}\bigg(\text{span}\bigg(\big\{ \expect_{\tilde{\bm{X}}^e}[\tilde{\bm{X}}^e (\tilde{\bm{X}}^e)^\top]X - \expect_{\tilde{\bm{X}}^e, \tilde{\epsilon}_y}[\tilde{\bm{X}}^e \tilde{\epsilon}_y]\big\}_{e \in \mathcal{E}_{partition}} \bigg) \bigg)\\
&\quad > d - r.
\end{align*}
\end{assumption}
\noindent Intuitively, this assumption states that we require the training environments in our partition partition $\mathcal{E}_{tr}$ to be sufficiently diverse, with limited co-linearity. 

\noindent Next, recall that in our setup, instead of directly observing the training partition for a given level $i$ and feature weight $v$, we need to identify the a subset $\mathcal{E}_{partition}$ from the available set of training environments $\mathcal{E}_{tr}$. Thus, for the training partition to be at least of size $m$, we need certain conditions on $|\mathcal{E}_{tr}|$. We characterize this in the following result.

\begin{lemma}
\label{lm:2}
%Under environment sampling as per equation \eqref{eq:gen_model}, if the cardinality of observed environments $|\mathcal{E}_{tr}| = n \sim  \frac{1}{\gamma} m \log(1/\delta)$, for partition that satisfies $\mathcal{E}_{partition} = \{e \in \mathcal{E}_{obs}: w_{i, e} = v\}$, we have that $P(|\mathcal{E}_{tr}| \geq m) > 1-\delta$. 
Under environment sampling as per equation \eqref{eq:gen_model}, if the cardinality of observed environments $|\mathcal{E}_{tr}| = n \sim  \frac{1}{\gamma} m \log(1/\epsilon)$, for the subset that satisfies the oracle distance condition i.e. $\mathcal{E}_{partition} := \{e \in \mathcal{E}_{tr}| \omega(e) = 1\}$, we have that $P(|\mathcal{E}_{partition}| \geq m) > 1-\epsilon$. 
\end{lemma}

\noindent Having obtained the correct partitioning with high probability, we call upon the out of domain generalization result from \cite{arjovskyBGP2020}.

\begin{proposition}[Theorem 9 in \cite{arjovskyBGP2020}]
\label{prop:1}
Assume that
\begin{equation}
\begin{gathered}
    Y^e = (Z_1^e)^\top \beta + \epsilon^e, \epsilon^e \perp Z_1^e, \  \expect[{\epsilon}^e] = 0\\
    X^e = S(Z_1^e, Z_2^e).
\end{gathered}
\end{equation}

Here, $\beta \in \mathbb{R}^c$, $Z_1^e$ takes values in $\mathbb{R}^c$, $Z_2^e$ takes values in $\mathbb{R}^1$ and $S \in \mathbb{R}^{d\times(c+q)}$. Assume that the $Z_1$ component of $S$ is invertible: that there exists $\tilde{S} \in \mathbb{R}^{c \times d}$ such that $\tilde{S}(S(z_1, z_2)) = z_1$, for all $z_1 \in \mathbb{R}^c$, $z_2 \in \mathbb{R}^q$. Let $\Phi \in \mathbb{R}^{d \times d}$ have rank $r$. Then, if atleast $d - r + d/r$ training environments $\mathcal{E}_{tr} \subseteq \mathcal{E}_{all}$ lie in linear general position of degree $r$, then we have:
$$\Phi \expect_{{X}^e}[{X}^e ({X}^e)^\top] \Phi^\top w = \Phi \expect_{{X}^e, Y^e}[{X}^e Y^e],$$
holds for $\forall e \in \mathcal{E}_{tr}$ iff $\Phi$ elicits an invariant predictor $\Phi^\top w$ for all $\ e \in \mathcal{E}_{all}$.
\end{proposition}

\noindent We now provide the proof of Theorem \ref{thm:2}. First, let $m = d - r + d/r$. Then from Lemma $\ref{lm:2}$, we know that sampling $n \sim \frac{1}{\gamma} m \log(1/\delta)$ environments gives us $|\mathcal{E}_{partition}| \geq m$ with high probability. From Assumption $3$, we also have that environments in $\mathcal{E}_{partition}$ lie in linear general position of degree $r$. Finally, note that our scheme is contingent on the oracle correctly identifying the required partition, which happens with probability greater than $(\frac{p}{p+1})^{|\mathcal{E}_{partition}|}$, as noted in Theorem 1. Armed with these, we can apply Proposition $\ref{prop:1}$ to our learning setup in equation \eqref{eq:observe_model} to learn an predictor $\bm{\Phi}, W$ that can recover features ${\bm{X}}_{inv}^e$ and corresponding desired weights $W_{inv}^e$ $\forall e \in \mathcal{E}_{all}$ which satisfy $w^{e}_i = w^{e_{ref}}_i$.

\clearpage

% \subsection{Proof of Lemma 2}

% \begin{itemize}
%      \item Given environment $e$, we assume that our contextual information $c(e)$ allows us to accurately predict the right partition i.e. both $P_{detection}(e \in \hat{\mathcal{E}}_{i, tr}|e \in {\mathcal{E}}_{i, tr}, c(e)) = p$ and $P_{detection}(e \notin \hat{\mathcal{E}}_{i, tr}|e \notin {\mathcal{E}}_{i, tr}, c(e)) = q$ is high.
%      \item Under the uniform feature model, we observe a set of random independent samples $\mathcal{E}_{tr}$, such that $|\mathcal{E}_{tr}| = n$.
%      \item Having sampled sufficient number of samples, we also need partition them correctly. When $M > k$, the probability of identifying the partition $\mathcal{E}_{tr, v} := \{e \in \mathcal{E}_{tr}: w_{i, e} = v\}$ by using prior $P(w_{i, e} = v|c(e))$ is lower bounded as $p^k(1-p)^{n-k}$.
%      \item Thus we also want the probability of misidentification to be low:
%      $$P(\hat{\mathcal{E}}_{tr, v} \neq {\mathcal{E}}_{tr, v}|M > m, p) < 1- p^m q^{n-m} < \delta_2$$
%      \item Suppose $\delta = \delta_1 + \delta_2$. Then by applying the union bound, we have that $P[(M > m) \cap (\mathcal{E}_{tr, v} = \hat{\mathcal{E}}_{tr, v})] > 1 - \delta$, which implies that the IRM predictor trained on environments in $\hat{\mathcal{E}}_{tr, v}$ will capture the feature weight $w_i =  v$ with high probability. 
% \end{itemize}
\newpage
\section{Additional Experiments}

In all our experiments when implementing IRM/P-IRM, we keep the penalty parameter sufficiently high $\lambda = 10^2/10^3$. The rationale for this is to have $\lambda$ high enough so that the invariance penalty term dominates the fidelity loss term and features are close to invariant. 
\label{sec:add-experiments}

\subsection{Synthetic Experiment}

The experimental setting is adapted from \cite{arjovskyBGP2020}. We assume the following generative model:
\begin{equation*}
    %\vspace{-1mm}
    \begin{gathered}
     \bm{X}_1 \leftarrow N(\bm{0}, e^2 I), \bm{X}_2 \leftarrow N(\bm{0}, e^2 I), \\
     %c(e) \leftarrow \text{Unif}(\{1, -1\}), \epsilon \sim N(0, \sigma(e)^2)\\
     {c(e) \in \{0, 1\}, \epsilon \sim N(\bm{0}, e^2 I)}\\
     \bm{y} \leftarrow \bm{X}_1^T W_1 + \bm{X}_2 ^T(c(e)W_2) + \epsilon,\ \epsilon \perp \bm{X}_1, \bm{X}_2\\
     %\bm{x}_3 \leftarrow \bm{y} + N(0, 1), \sigma(e)^2 \in [0,  \sigma_{MAX}^2].\\
    \end{gathered}
\end{equation*}

\begin{figure}[!ht]
    \centering
    \includegraphics[width=.7\linewidth]{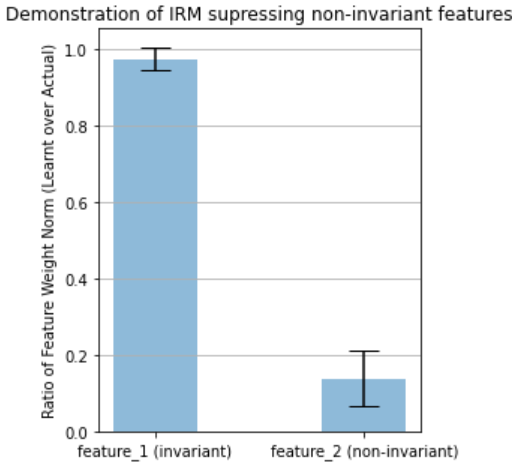}
    \caption{The plot demonstrates that IRM is incentivized to suppress non-invariant features, as is the case for feature\_2.}
    \label{fig:synthetic}
\end{figure}

\noindent The task is to predict target $\bm{y} \in \mathbb{R}$ based on observed $\bm{X} \in \mathbb{R}^{20}$, where $\bm{X}=(\bm{X}_1 \in \mathbb{R}^{10}, \bm{X}_2 \in \mathbb{R}^{10})$. The predictor $w \circ \Phi$ is learnt via the IRM objective and as noted previously, the overparametrization in the objective is handled by parametrizing $\Phi \in \mathbb{R}^{20}$ to be rank $1$ and fixing $w=1.0$ as a scalar. The true weights $W_1, W_2 \in \mathbb{R}^{10}$ are fixed Gaussian entries, but the sampling of $c(e)$ for different environments controls whether $\bm{X}_2$ is a causal feature for $\bm{y}$.\\
The goal is to visualize the intuition in Lemma~\ref{lemma:1}  i.e. how IRM can discard causal features which are non-invariant. To that end, we sample $1000$ data points from four environments characterized by $e \in \{0.2, 1, 2, 5\}$. The $c(e)$ for each environment is assigned uniformly, such that the final training set has two environments for each of $\{0, 1\}$, and feature $\bm{X}_2$ is non-invariant. The learning procedure for IRM is consistent with the rest of the paper, with $\lambda = 10^3$ and the initial $4000$ epochs for annealing the IRM loss.  \\
To study learning of each feature, first denote $\Phi = (\Phi_1, \Phi_2)$, $\Phi_1, \Phi_2 \in \mathbb{R}^{10}$. Then note that $\Phi_i$ captures the contribution of feature $\bm{X}_i$ in the prediction. We then look at $\frac{\|\Phi_i\|}{\|W_i\|}$ (averaged over the random draw of environment weights). Intuitively, this ratio indicates the information captured by the IRM predictor for that feature. We visualize the results in Fig.~\ref{fig:synthetic}, which demonstrates the tendency for IRM to suppress learning of non-invariant features.  

\subsection{Linear Regression}

For our choice of learning rate, number of iterations and optimizer and annealing iterations, we refer to (\cite{lin2022zin}). While the reported results were for $\lambda = 10^2$, we verified similar trends for $\lambda = 10^3$.

  % Table generated by Excel2LaTeX from sheet 'params table Metashift'
  \begin{table}[!ht]
    \centering
      \begin{tabular}{ll}
      \toprule
      Hyperparameter & Values \\
      \midrule
      Number of Iterations & 4000 \\
      Learning rate & $10^{-3}$ \\
      Optimizer & Adam \\
      IRM Penalty & $10^{2}$ \\
      Annealing Iterations & 2000 \\
      \bottomrule
      \end{tabular}%
      \vspace{6pt}
    \caption{Hyperparameters for experiments on the housing dataset, following \cite{lin2022zin}}.
    \label{tab:apx_params_housing}%
  \end{table}%

\begin{figure}[!ht]
    \centering
    \includegraphics[width=.9\linewidth]{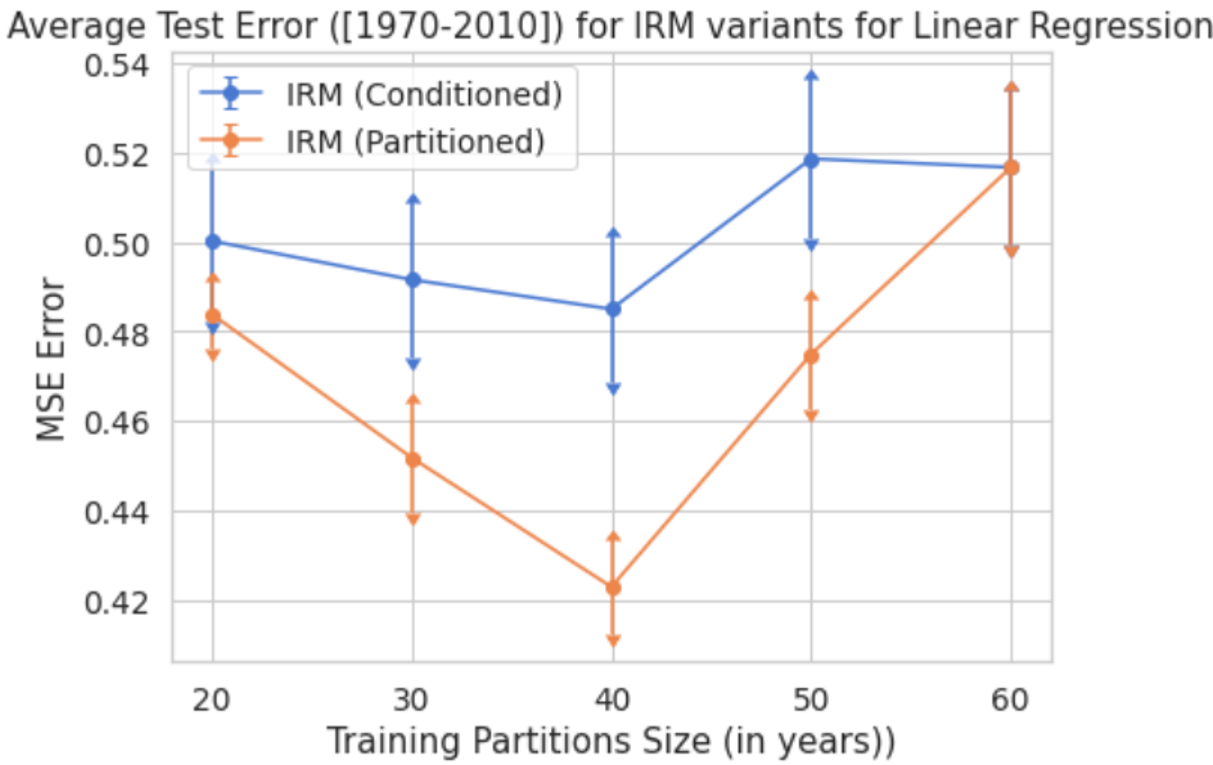}
    \caption{\textbf{Partitioning vs Conditioning}: In the underparametrized linear regression setting where the number of data points is much greater than learnable parameters, conditioning is not helpful in terms of improving P-IRM accuracy and partitioning consistently performs better.}
    \label{fig:housing_cond_vs_part}
\end{figure}

\begin{figure}[!ht]
    \centering
    \includegraphics[width=.9\linewidth]{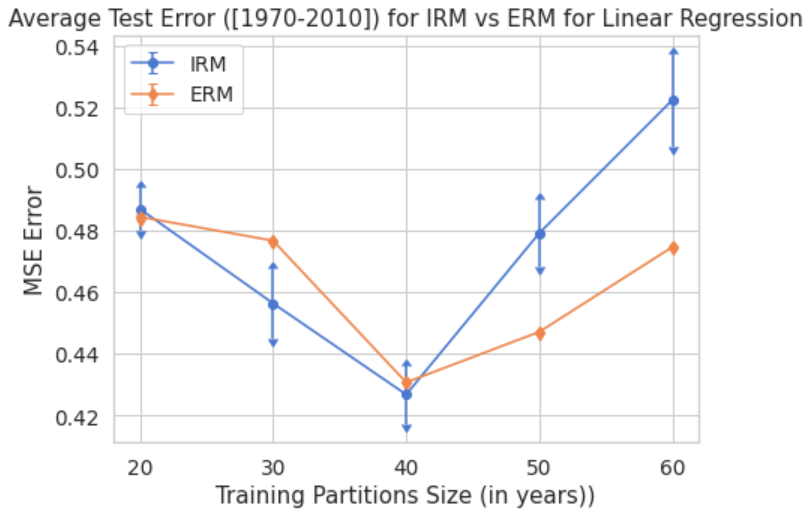}
    \caption{For the regression experiment in the main paper, we find that for both ERM and IRM, there exists an optimal partition. Note that while ERM consistently finds the unique optima, the IRM solution has some variance due to the non-convex objective.}
    \label{fig:housing_erm_vs_irm}
\end{figure}
\subsection{Image Classification}
For the image classification experiments on MetaShift \cite{liang2022metashift} dataset, we follow a similar training pipeline as in \cite{gulrajaniL2020,liang2022metashift}. Following \cite{gulrajaniL2020}, we consider ResNet-50 \cite{he2016resnet}, since larger ResNets are known to generalize better. ResNet-50 was pre-trained on ImageNet \cite{he2016resnet}, and for domain generalization, the batch normalization and final softmax layers of ResNet are chopped off \cite{gulrajaniL2020}. Then ResNet-50 layers are followed by non-linear functions, e.g. ReLU, and a final dropout layer \cite{gulrajaniL2020}. 
\subsubsection{Domain Generalization}
\label{apx:domgen}
  % Table generated by Excel2LaTeX from sheet 'better_looking'
  \begin{table*}[!ht]
  \small
    \centering
    \tabcolsep=0.03cm
      \begin{tabular}{cccccc}
      \toprule
          &
          \multicolumn{1}{l}{Experiment 1} & \multicolumn{1}{l}{Experiment 2} & \multicolumn{1}{l}{Experiment 3} & \multicolumn{1}{l}{Experiment 4} &  \\
      \midrule
          & $d=0.17$ & $d=0.54$ & $d=0.81$ & $d=0.92$ & Avg. Performance \\
      \midrule
      P-ERM $(p=25)$ & \textbf{0.867(0.050)} & 0.740(0.079) & \textbf{0.557(0.056)} & 0.430(0.079) & 0.6485 \\
      \midrule
      P-IRM $(p=25)$ & 0.820(0.148) & 0.742(0.138) & 0.597(0.243) & \textbf{0.753(0.209)} & 0.728 \\
      \midrule
      P-IB\_IRM $(p=25)$ & 0.613(0.386) & 0.740(0.171) & 0.203(0.029) & 0.343(0.464) & 0.47475 \\
      \bottomrule
      \end{tabular}%
      \normalsize	
      % \vspace{-6pt}
     \caption{Continued table on Domaing Generalization in Metashift, for $p = 25$}
    \label{tab:metashift_domain_generalization_DG_p_25}%
    % \vspace{-6pt}
  \end{table*}%

Herein, we report the relevant choice of hyperparameters in Table \ref{tab:metashift_dg_params_irm},  to reproduce our results pertaining to the domain generalization experiment in the main paper. 
 % Table generated by Excel2LaTeX from sheet 'dg'
 \begin{table*}[htp]
   \centering
   \small
    \tabcolsep=0.001cm
  % \scalebox{0.68}{
     \begin{tabular}{ccccc}
     \toprule
     \multicolumn{1}{p{9.165em}}{Model} & \multicolumn{1}{p{5em}}{IRM Penalty weight} & \multicolumn{1}{p{5em}}{IRM Annealing Iterations} & \multicolumn{1}{p{5em}}{IB\_IRM Penalty weight} & \multicolumn{1}{p{5em}}{IB\_IRM annealing iterations } \\
     \midrule
     Experiment 1 &   &   &   &  \\
     \midrule
     IRM & 10 & 20 &   &  \\
     P-IRM (p=0) & 10 & 20 &   &  \\
     P-IRM (p=10) & 10 & 40 &   &  \\
     P-IRM (p=25) & 10 & 20 &   &  \\
     \midrule
     IB\_IRM & 10 & 40 & 10 & 20 \\
     P-IB\_IRM (p=0) & 10 & 20 & 10 & 40 \\
     P-IB\_IRM (p=10) & 100 & 20 & 10 & 20 \\
     P-IB\_IRM (p=25) & 10 & 40 & 10 & 20 \\
       &   &   &   &  \\
     \midrule
     Experiment 2 &   &   &   &  \\
     \midrule
     IRM & 10 & 20 &   &  \\
     P-IRM (p=0) & 10 & 20 &   &  \\
     P-IRM (p=10) & 10 & 20 &   &  \\
     P-IRM (p=25) & 10 & 20 &   &  \\
     \midrule
     IB\_IRM & 10 & 20 & 10 & 20 \\
     P-IB\_IRM (p=0) & 10 & 20 & 10 & 20 \\
     P-IB\_IRM (p=10) & 10 & 20 & 10 & 20 \\
     P-IB\_IRM (p=25) & 10 & 20 & 10 & 20 \\
       &   &   &   &  \\
     \midrule
     Experiment 3 &   &   &   &  \\
     \midrule
     IRM & 10 & 40 &   &  \\
     P-IRM (p=0) & 10 & 40 &   &  \\
     P-IRM (p=10) & 10 & 40 &   &  \\
     P-IRM (p=25) & 10 & 40 &   &  \\
     \midrule
     IB\_IRM & 10 & 20 & 10 & 20 \\
     P-IB\_IRM (p=0) & 10 & 20 & 10 & 40 \\
     P-IB\_IRM (p=10) & 10 & 20 & 10 & 20 \\
     P-IB\_IRM (p=25) & 10 & 20 & 10 & 20 \\
       &   &   &   &  \\
     \midrule
     Experiment 4 &   &   &   &  \\
     \midrule
     IRM & 10 & 20 &   &  \\
     P-IRM (p=0) & 10 & 40 &   &  \\
     P-IRM (p=10) & 10 & 20 &   &  \\
     P-IRM (p=25) & 10 & 20 &   &  \\
     \midrule
     IB\_IRM & 10 & 20 & 10 & 20 \\
     P-IB\_IRM (p=0) & 10 & 20 & 10 & 20 \\
     P-IB\_IRM (p=10) & 10 & 40 & 10 & 20 \\
     P-IB\_IRM (p=25) & 10 & 40 & 10 & 20 \\
     \bottomrule
     \end{tabular}%
     % }
     \caption{Choice of hyper-parameters for Domain Generalization experiments for IRM and IB\_IRM,  obtained via train-domain validation.}
   \label{tab:metashift_dg_params_irm}%
 \end{table*}%

\subsubsection{Subpopulation Shift}
\label{apx:subshifts}
Following our setup described in %~\nameref{sec:metashift_section}
main paper, we conduct further experiments to study performance under \textit{subpopulation shifts} for the binary classification task. In subpopulation shifts, communities used for training and testing are the same, but their relative proportions differ between training and testing environments, with certain groups often subject to under-representation. The goal is to obtain a model to do well even for minority groups in the training data \cite{koh_wilds2019}. \\
Following \cite{liang2022metashift}, the communities are grouped into two environments: indoor and outdoor. In training,  $cat(outdoor)$ and $dog(indoor)$ subsets are the minority groups, while $cat(indoor)$ and $dog(outdoor)$ are majority groups.  We vary the percentage of minority groups within the training set to be $m \in \{0.12, 0.01\}$ of the total training set, and we keep the size of the training set fixed with 1700 samples. We use a balanced set of testing by equally sampling from each environment with balanced labels.   
Tables ~\ref{tab:subpopulation} and ~\ref{tab:subpopulation2} show that the results under subpopulation shift settings. We report the average accuracy over the four groups, and worst group accuracy (the group with the worst performace), and the average minority accuracy which is the average of minority groups in training i.e. $cat(outdoor)$ and $dog(indoor)$. On the more challenging setting of $m=0.01$ where minority groups are observed minimally, P-IRM models achieve better worst group and average minority performance. However as expected, it becomes harder to improve performance for partitioned models as we lower the amount of available training data, and it is best rely on IRM/ERM.

\noindent For reproducibility of results, Table \ref{tab:irm_params_sps} shows the selected hyperparameters. 
  % Table generated by Excel2LaTeX from sheet 'Sheet5'
  \begin{table*}[ht]
    \centering
    \small
     \tabcolsep=0.5cm
        % \scalebox{0.75}{
      \begin{tabular}{l|lll}
      \toprule
          & $m=0.12$  &     &  \\
      \midrule
          & Avg. Acc. & Worst Group Acc.  & Avg. Minority Acc.  \\
      \midrule
      ERM & \textbf{0.816(0.209)} & \textbf{0.722(0.103)} & \textbf{0.737(0.021)} \\
      P-ERM $(p=0)$ & 0.78(0.129) & 0.623(0.157) & 0.675(0.074) \\
      P-ERM  $(p=10)$ & 0.76(0.189) & 0.526(0.089) & 0.616(0.127) \\
      P-ERM $(p=25)$ & 0.779(0.154) & 0.590(0.043) & 0.736(0.206) \\
      \midrule
      IRM & 0.638(0.250) & 0.336(0.160) & 0.558(0.314) \\
      P-IRM $(p=0)$ & 0.703(0.158) & 0.560(0.230) & 0.5705(0.015) \\
      P-IRM  $(p=10)$ & 0.681(0.190) & 0.518(0.143) & 0.663(0.205) \\
      P-IRM $(p=25)$ & \textbf{0.737(0.147)} & \textbf{0.604(0.227)} & \textbf{0.7335(0.183)} \\
      \midrule
      IB\_IRM & \textbf{0.639(0.209)} & 0.380(0.082) & 0.47(0.127) \\
      P-IB\_IRM $(p=0)$ & 0.587(0.136) & \textbf{0.451(0.234)} & 0.4695(0.026) \\
      P-IB\_IRM $(p=10)$ & 0.613(0.301) & 0.264(0.137) & \textbf{0.5635(0.424)} \\
      P-IB\_IRM $(p=25)$ & 0.596(0.178) & 0.426(0.115) & 0.448(0.031) \\
      \bottomrule
      \end{tabular}
      % }
    \caption{Subpopulation shift on Metashift: The value $m$ represents the portion of minority groups within a training environment. The partitioned models are applied with additional samples up to percentage $p \in \{0, 10, 25\}$}
    \label{tab:subpopulation}%
  \end{table*}%

  % Table generated by Excel2LaTeX from sheet 'Sheet5'
  \begin{table*}[ht]
    \centering
    \small
     \tabcolsep=0.5cm
  % \scalebox{0.75}{
      \begin{tabular}{l|lll}
      \toprule
          & $m=0.01$ &     &  \\
      \midrule
          & Avg. Acc. & Worst Group Acc.  & Avg. Minority Acc.  \\
      \midrule
      ERM & 0.744(0.209) & 0.514(0.113) & 0.559(0.064) \\
      P-ERM $(p=0)$ & \textbf{0.747(0.186)} & \textbf{0.574(0.112)} & \textbf{0.587(0.018)} \\
      P-ERM  $(p=10)$ & 0.729(0.254) & 0.481(0.133) & 0.51(0.041) \\
      P-ERM $(p=25)$ & 0.745(0.248) & 0.488(0.013) & 0.532(0.062) \\
      \midrule
      IRM & \textbf{0.725(0.185)} & 0.509(0.216) & 0.5775(0.097) \\
      P-IRM $(p=0)$ & 0.677(0.232) & 0.426(0.116) & 0.484(0.082) \\
      P-IRM  $(p=10)$ & 0.629(0.277) & 0.341(0.260) & \textbf{0.6355(0.269)} \\
      P-IRM $(p=25)$ & 0.687(0.181) & \textbf{0.525(0.189)} & 0.531(0.008) \\
      \midrule
      IB\_IRM & 0.269(0.179) & \textbf{0.470(0.379)} & 0.5555(0.121) \\
      P-IB\_IRM $(p=0)$ & 0.56(0.27) & 0.208(0.102) & 0.3955(0.265) \\
      P-IB\_IRM $(p=10)$ & 0.556(0.206) & 0.366(0.108) & 0.4805(0.162) \\
      P-IB\_IRM $(p=25)$ & \textbf{0.568(0.138)} & 0.398(0.412) & \textbf{0.583(0.092)} \\
      \bottomrule
      \end{tabular}%
      % }
    \caption{Subpopulation shift: The value $m$ represents the portion of minority groups within a training environment. The partitioned models are applied with additional samples up to percentage $p \in {0, 10, 25}$}
    \label{tab:subpopulation2}%
  \end{table*}%

 % Table generated by Excel2LaTeX from sheet 'spb'
 \begin{table*}[!ht]
   \centering
   \small
   \tabcolsep=0.01cm
  % \scalebox{0.68}{
     \begin{tabular}{c|cccc}
     \toprule
     \multicolumn{1}{p{11em}}{Model} & \multicolumn{1}{p{5em}}{IRM Penalty weight} & \multicolumn{1}{p{5em}}{IRM Annealing Iterations} & \multicolumn{1}{p{5em}}{IB\_IRM Penalty weight} & \multicolumn{1}{p{5em}}{IB\_IRM annealing iterations } \\
     \midrule
     \multicolumn{1}{c}{Experiment 1 ($m=0.12$ )} &   &   &   &  \\
     \midrule
     IRM & 10 & 40 &   &  \\
     P-IRM $(p=0)$ & 10 & 40 | 20 &   &  \\
     P-IRM  $(p=10)$ & 10 & 40 | 20 &   &  \\
     P-IRM $(p=25)$ & 10 & 20 | 40 &   &  \\
     \midrule
     IB\_IRM & 10 & 40 & 10 & 20 \\
     P-IB\_IRM $(p=0)$ & 10 & 20 & 10 & 20 \\
     P-IB\_IRM $(p=10)$ & 10 & 20 | 40 & 10 & 40 \\
     P-IB\_IRM $(p=25)$ & 10 & 20 & 10 & 20 | 40 \\
     \midrule
     \multicolumn{1}{c}{} &   &   &   &  \\
     \midrule
     \multicolumn{1}{c}{Experiment 2 ($m=0.01$ )} &   &   &   &  \\
     \midrule
     IRM & 10 & 40 &   &  \\
     P-IRM $(p=0)$ & 1000 | 10 & 40 &   &  \\
     P-IRM  $(p=10)$ & 100 | 10 & 40 | 20 &   &  \\
     P-IRM $(p=25)$ & 100 | 10 & 20 | 40 &   &  \\
     \midrule
     IB\_IRM & 1000 & 40 & 10 & 40 \\
     P-IB\_IRM $(p=0)$ & 10 & 40 | 20 & 100 | 10 & 20 \\
     P-IB\_IRM $(p=10)$ & 100 | 10 & 20 & 100 | 10 & 40 | 20 \\
     P-IB\_IRM $(p=25)$ & 1000 | 10 & 20 & 100 | 10 & 40 | 20 \\
     \bottomrule
     \end{tabular}
     % }
     \caption{Choice of hyper-parameters for Subpopulation Shifts experiment on Metashift for IRM and IB\_IRM,  obtained via train-domain validation. (For partitioned models since we are using two models, we are reporting the parameters for both, if they are the same we report one value only)}
   \label{tab:irm_params_sps}
 \end{table*}%

\clearpage
\subsection{Language Experiments}
For language experiments, NER and TC, we build a classifier based on the pre-trained language model BERT \cite{devlin2019bert}, followed by a dropout and a linear layer. We also considered DistillBERT \cite{sanh2019distilbert} and GPT-2 \cite{radford2019language},but found that BERT-based models outperformed other networks. 
We train the models for the maximum number of iterations, (details in tables \ref{tab:apx_params_language}, and \ref{tab:apx_params_aic}) for one seed. Then we select the best number of iterations to apply for other seeds. The results are the average of three seeds. 
\subsubsection{Named Entity Recognition (NER)}

For the language NER experiments, the best hyperparameter values are reported in Table~\ref{tab:apx_params_language}, and Table~\ref{tab:irm_params} which have been selected based on the best model performance on the validation set. The training was done for 80 epochs, around which both training and in-domain validation losses stabilize and remain the same. For annealing epochs, we considered [10, 20, 30, 35, 40] epochs and found that for all variants of P-IRM/IRM, 40 epochs yielded best performance. The optimizer and learning rate was based on standard choice for using pre-trained BERT models. 

  % Table generated by Excel2LaTeX from sheet 'params table Metashift'
  \begin{table}[htbp]
    \centering
      \begin{tabular}{ll}
      \toprule
      Hyperparameter & Values \\
      \midrule
      Maximum Number of epochs & 80 \\
      Batch size & 8 \\
      Learning rate & $10^{-6}$ \\
      Optimizer & Adam \\
Number of GPUs & 4 \\
      \bottomrule
      \end{tabular}%
    \caption{Hyperparameters choices for experiments on the NER dataset}
    \vspace{-1em}
    \label{tab:apx_params_language}%
  \end{table}%

% Finally, the choice of batch size set to 8. We note that choice of batch size and the corresponding stochasticity in gradient descent can indeed affect IRM performance to a significant degree. Indeed, when we reduce batch size to two, while we still find P-IRM variants marginally better than IRM, the IRM overall accuracy is much worse than ERM, which is not affected to the same degree. 

\subsubsection{Text Classification (TC)}
\label{apx:aic}
We consider another language classification task, which identifies the venue of a published paper \footnote{https://www.semanticscholar.org/product/api}, selecting AAAI and ICML conferences for classification. This task represents a topic classification task. %or author disambiguation. 
Our temporal partitions are: {2006-2008, 2009-2011, 2012-2014, and 2015-2017} and we test on papers published between 2018-2020. The model selection follows as before, training for a maximum of 40 epochs. The final hyperparameters are reported in Table \ref{tab:irm_params_aic} and \ref{tab:apx_params_aic}. Table \ref{tab:apx_aic_results} shows how partitioning outperforms their baseline methods. P-IRM with two environments performed the best among all other models. 
  % Table generated by Excel2LaTeX from sheet 'params table Metashift'
  \begin{table}[htbp]
    \centering
      \begin{tabular}{ll}
      \toprule
      Hyperparameter & Values \\
      \midrule
      Maximum Number of epochs & 40 \\
      Batch size & 8 \\
      Learning rate & $10^{-6}$ \\
      Optimizer & Adam \\
Number of GPUs & 4 \\
      \bottomrule
      \end{tabular}%
    \caption{Hyperparameters choices for the Text Classification task}
    \vspace{-1em}
    \label{tab:apx_params_aic}%
  \end{table}%

%   % Table generated by Excel2LaTeX from sheet 'Sheet1'
%   \begin{table}[!ht]
%     \centering
%     \tabcolsep=0.01cm
%   \scalebox{0.65}{
%      \begin{tabular}{p{10.165em}rrrrr}
%      \toprule
%      Model & \multicolumn{1}{p{3.75em}}{\# envs} & \multicolumn{1}{p{4.75em}}{IRM Penalty weight} & \multicolumn{1}{p{6.165em}}{IRM Annealing Iterations} & \multicolumn{1}{p{6.165em}}{IB\_IRM Penalty weight} & \multicolumn{1}{p{6.335em}}{IB\_IRM annealing iterations } \\
%      \midrule
%      IRM & 4 & 1000 & 30 &   &  \\
%      P-IRM (partitioned) & 3 & 1000 & 40 &   &  \\
%      P-IRM (partitioned) & 2 & 1000 & 40 &   &  \\
%      P-IRM (conditioned) & 3 & 100 & 40 &   &  \\
%      P-IRM (conditioned) & 2 & 100 & 30 &   &  \\
%      \midrule
%      IB\_IRM & 4 & 100 & 40 & 1 & 40 \\a
%      P-IB\_IRM (partitioned) & 3 & 100 & 40 & 1 & 40 \\
%      P-IB\_IRM (partitioned) & 2 & 100 & 40 & 1 & 40 \\
%      P-IB\_IRM (conditioned) & 3 & 100 & 40 & 1 & 40 \\
%      P-IB\_IRM (conditioned) & 2 & 100 & 40 & 1 & 40 \\
%      \bottomrule
%      \end{tabular}%
% }
%       \caption{NER dataset: Best IRM hyperparameters values selected based on early stopping on validation data}
%     \label{tab:irm_params}%
%   \end{table}%

 % Table generated by Excel2LaTeX from sheet 'scierc'
 \begin{table*}[!ht]
   \centering
   \small
   \tabcolsep=0.06cm
    % \scalebox{0.60}{
     \begin{tabular}{p{10.165em}llllll}
     \toprule
     Model & \multicolumn{1}{p{3.75em}}{\# envs} & \multicolumn{1}{p{4.75em}}{IRM Penalty weight} & \multicolumn{1}{p{4.465em}}{IRM annealing iterations} & \multicolumn{1}{p{5.165em}}{IB\_IRM Penalty weight} & \multicolumn{1}{p{5.335em}}{IB\_IRM annealing iterations } & \multicolumn{1}{p{5em}}{\#epochs} \\
     \midrule
     ERM & 4 &   &   &   &   & 44 \\
     P-ERM & 3 &   &   &   &   & 54 \\
     P-ERM & 4 &   &   &   &   & 58 \\
     \midrule
     IRM & 4 & 1000 & 30 &   &   & 53 \\
     P-IRM (partitioned) & 3 & 1000 & 40 &   &   & 70 \\
     P-IRM (partitioned) & 2 & 1000 & 40 &   &   & 76 \\
     P-IRM (conditioned) & 3 & 100 & 40 &   &   & 64 \\
     P-IRM (conditioned) & 2 & 100 & 30 &   &   & 66 \\
     \midrule
     IB\_IRM & 4 & 100 & 40 & 1 & 40 & 57 \\
     P-IB\_IRM (partitioned) & 3 & 100 & 40 & 1 & 40 & 77 \\
     P-IB\_IRM (partitioned) & 2 & 100 & 40 & 1 & 40 & 76 \\
     P-IB\_IRM (conditioned) & 3 & 100 & 40 & 1 & 40 & 76 \\
     P-IB\_IRM (conditioned) & 2 & 100 & 40 & 1 & 40 & 59 \\
     \bottomrule
     \end{tabular}%
     % }
   \caption{NER dataset: Best IRM hyperparameters values selected based on early stopping on validation data}
    \label{tab:irm_params}%
 \end{table*}%

  \begin{table*}[!ht]
  \small
    \centering
    \tabcolsep=0.2cm
  % \scalebox{0.70}{
      \begin{tabular}{cccc}
      \toprule
      Model & \multicolumn{1}{p{5em}}{Number of envs} & \multicolumn{1}{p{5em}}{Training Years} & \multicolumn{1}{p{12.335em}}{Testing accuracy (2018-2020)} \\
      \midrule
      ERM & 4   & 2006-2017 & 0.862 (0.008) \\
      P-ERM & 3   & 2009-2017 & 0.862 (0.004) \\
      P-ERM & 2   & 2012-2017 & \textbf{0.875 (0.014)} \\
      \midrule
      IRM & 4   & 2006-2017 & 0.846 (0.013) \\
      P-IRM (paritioned) & 3   & 2009-2017 & 0.862 (0.008) \\
      P-IRM (paritioned) & 2   & 2012-2017 & \textbf{0.882 (0.016)} \\
      P-IRM (conditioned) & 3   & 2009-2017 & 0.869 (0.007) \\
      P-IRM (conditioned) & 2   & 2012-2017 & 0.853 (0.010) \\
      \midrule
      IB\_IRM & 4   & 2006-2017 & 0.846 (0.014) \\
      P-IB\_IRM (partitioned) & 3   & 2009-2017 & 0.868 (0.001) \\
      P-IB\_IRM (partitioned) & 2   & 2012-2017 & \textbf{0.874 (0.016)} \\
      P-IB\_IRM (conditioned) & 3   & 2009-2017 & 0.860 (0.016) \\
      P-IB\_IRM (conditioned) & 2   & 2012-2017 & 0.862 (0.011) \\
      \bottomrule
      \end{tabular}%
      % }
    \caption{TC dataset: Results on text classification, comparison between ERM, IRM, IB\_IRM and their partitioned variants. }
    \label{tab:apx_aic_results}%
  \end{table*}%

 % Table generated by Excel2LaTeX from sheet 'aic'
\begin{table*}[!ht]
   \centering
   \small
   \tabcolsep=0.06cm
    % \scalebox{0.60}{
     \begin{tabular}{p{5.585em}llllll}
     \toprule
     \multicolumn{1}{l}{Model} & \multicolumn{1}{p{5em}}{\# envs} & \multicolumn{1}{p{5em}}{IRM Penalty weight} & \multicolumn{1}{p{5em}}{IRM Annealing Iterations} & \multicolumn{1}{p{5em}}{IB\_IRM Penalty weight} & \multicolumn{1}{p{5em}}{IB\_IRM annealing iterations } & \multicolumn{1}{p{5em}}{\# epochs} \\
     \midrule
     ERM & 4 &   &   &   &   & 22 \\
     P-ERM & 3 &   &   &   &   & 39 \\
     P-ERM & 4 &   &   &   &   & 38 \\
     \midrule
     IRM & 4 & 1000 & 20 &   &   & 37 \\
     P-IRM (partitioned) & 3 & 1000 & 20 &   &   & 36 \\
     P-IRM (partitioned) & 2 & 1000 & 20 &   &   & 37 \\
     P-IRM (conditioned) & 3 & 1000 & 20 &   &   & 33 \\
     P-IRM (conditioned) & 2 & 1000 & 20 &   &   & 33 \\
     \midrule
     IB\_IRM & 4 & 1000 & 20 & 0.1 & 20 & 37 \\
     P-IB\_IRM (partitioned) & 3 & 1000 & 20 & 0.1 & 20 & 36 \\
     P-IB\_IRM (partitioned) & 2 & 1000 & 20 & 0.1 & 20 & 39 \\
     P-IB\_IRM (conditioned) & 3 & 1000 & 20 & 0.1 & 20 & 33 \\
     P-IB\_IRM (conditioned) & 2 & 1000 & 20 & 0.1 & 20 & 33 \\
     \bottomrule
     \end{tabular}%
     % }
     \caption{TC dataset: Best IRM hyperparameters values selected based on early stopping on validation data}
   \label{tab:irm_params_aic}%
 \end{table*}%

\end{document}